\documentclass[11pt]{article}

\usepackage[final]{acl}

\usepackage{times}
\usepackage{latexsym}

\usepackage[T1]{fontenc}

\usepackage[utf8]{inputenc}

\usepackage{microtype}

\usepackage{inconsolata}

\usepackage{graphicx}
\usepackage{microtype}
\usepackage{subcaption}
\usepackage{enumitem}
\usepackage{multirow}
\usepackage{booktabs} 
\usepackage{booktabs,threeparttable,xcolor,colortbl}
\usepackage{tcolorbox}
\usepackage{amsmath}
\usepackage{amssymb}
\usepackage{mathtools}
\usepackage{amsthm}
\usepackage[utf8]{inputenc}
\usepackage{microtype}
\usepackage{newunicodechar}
\newunicodechar{−}{\textminus}
\usepackage{algorithm}
\usepackage{algpseudocode}

\usepackage{tcolorbox}
\tcbuselibrary{breakable}

\definecolor{headerblue}{HTML}{A2C8E9}
\definecolor{datacellgreen}{HTML}{9DDDA5}   
\definecolor{bestoverall}{HTML}{F8CC87}
\newcommand{\best}[1]{\cellcolor{datacellgreen}\textbf{#1}}         
\newcommand{\overallbest}[1]{\cellcolor{bestoverall}\textbf{#1}}  

\definecolor{goodgreen}{RGB}{198,239,206}
\definecolor{neutralyellow}{RGB}{255,248,204}
\definecolor{badred}{RGB}{255,179,196}
\definecolor{lightorange}{RGB}{255,207,179}
\definecolor{warnyellow}{RGB}{255,238,200}

\title{
EvidenceRL: Reinforcing Evidence Consistency \\for Trustworthy Language Models
}

\author{
  J. Ben Tamo\thanks{~~Equal contribution.}\affA,
  Yuxing Lu\footnotemark[1]\affA \affB,
  Benoit L. Marteau\affA,
  Micky C. Nnamdi\affA,\\
  \and
  \textbf{May D. Wang \thanks{~~Corresponding author.}}\affA
  \\[4pt]
  \affA Georgia Institute of Technology \\
  \affB Pekin University
  \\[4pt]
  \texttt{\{jtamo3, yxlu, bmarteau3, mnnamdi3, maywang\}@gatech.edu }
  \\[6pt]
}

\newcommand{\affA}{\textsuperscript{$\spadesuit$}} 
\newcommand{\affB}{\textsuperscript{$\heartsuit$}} 

\begin{document}
\maketitle
\begin{abstract}
Large Language Models (LLMs) are fluent but prone to hallucinations, producing answers that appear plausible yet are unsupported by available evidence. This failure is especially problematic in high-stakes domains where decisions must be justified by verifiable information.
We introduce \textbf{EvidenceRL}, a reinforcement learning framework that enforces evidence adherence during training. EvidenceRL scores candidate responses for grounding (entailment with retrieved evidence and context) and correctness (agreement with reference answers) and optimizes the generator using Group Relative Policy Optimization (GRPO). 
We evaluate across two high-stakes domains, cardiac diagnosis and legal reasoning, where EvidenceRL consistently improves evidence grounding and faithfulness without sacrificing task accuracy. On cardiac diagnosis, F1@3 increases from 37.0 to 54.5 on Llama-3.2-3B while grounding ($G_{\max}@3$) rises from 47.6 to 78.2; hallucinations drop nearly 5$\times$ and evidence-supported diagnoses increase from 31.8\% to 61.6\%. On legal reasoning, EvidenceRL raises Faithfulness from 32.8\% to 67.6\% on Llama-3.1-8B, demonstrating consistent behavioral change across domains. Our code is open-sourced at \url{https://github.com/Wizaaard/EvidenceRL.git}.

\end{abstract}

\section{Introduction}
\label{sec:intro}
The integration of Large Language Models (LLMs) into safety-critical domains such as healthcare, law, and finance has exposed reliability failures that limit real-world deployment \cite{gallagher2024assessing}. 
A key driver of these failures is the lack of evidence grounding: models often produce plausible outputs that are not supported by the available evidence 
\cite{gallagher2024assessing,magesh2025hallucination,nanda2025state,sarvari2025challenges}.
These findings indicate that without mechanisms to enforce evidence adherence, LLMs remain unreliable for high-stakes decision support.

Retrieval-Augmented Generation (RAG) attempts to mitigate hallucinations by conditioning generation on retrieved documents \cite{lewis2020retrieval}.
However, retrieval alone does not guarantee that outputs are derived from the provided evidence, and models frequently produce unsupported answers even when relevant context is available \cite{gao2023retrieval,zhang-etal-2025-faithfulrag,10.1145/3731120.3744592}.
This failure mode, often termed context-memory conflict, persists even when high-quality evidence is provided, revealing a core limitation of existing RAG pipelines: retrieval is treated as a soft prompt rather than a constraint on generation.

Existing approaches address reliability through either inference-time filtering or training-time alignment. Post-hoc verification methods detect unsupported statements after generation \cite{manakul2023selfcheckgpt,farquhar2024detecting}, while alignment techniques such as instruction tuning and reinforcement learning from human feedback (RLHF) shape model behavior but do not explicitly optimize evidence grounding \cite{peng2023instruction,bai2022training}. Hybrid approaches introduce retrieval or verification during reasoning \cite{asai2023self,dhuliawala2024chain}, yet evidence usage remains loosely coupled to the training objective. As a result, current systems lack mechanisms that directly enforce evidence-consistent generation during learning.

To address these limitations, we introduce EvidenceRL, a reinforcement learning framework that optimizes evidence adherence during training rather than enforcing it post hoc. Unlike standard alignment methods that rely on sparse or noisy holistic rewards, EvidenceRL integrates two complementary signals: (1) a fine-grained grounding reward using a Focus-Then-Verify architecture to compute sentence-level entailment against the context and prevent signal dilution; and (2) a semantic correctness reward using an LLM judge to verify domain-specific answer equivalence. 
Across multiple model families (Llama, Gemma, GPT-oss) and scales (3B–120B), EvidenceRL consistently improves both task accuracy and evidential grounding. In the medical domain (MIMIC-IV-Ext), diagnostic F1@3 improves by up to 17 points while grounding increases substantially (e.g., on Llama-3.2-3B, $G_{\max}@3$ rises from 47.6 to 78.2). In the legal domain (BarExam MBE), EvidenceRL similarly shifts predictions toward evidence-supported reasoning, increasing the Evidence-Based rate from 18.8\% to 41.0\% and Faithfulness from 32.8\% to 67.6\% on Llama-3.1-8B. Behavioral analysis across both domains shows a clear redistribution toward evidence-based predictions: hallucinations drop by nearly $5\times$, while evidence-supported answers increase substantially. 

Our contributions are as follows:
\begin{itemize}
\item We propose \textbf{EvidenceRL}, a reinforcement learning framework that \textbf{enforces} evidence grounding as a differentiable training objective using automated NLI and LLM-judge rewards, removing the dependency on human preference annotations.
\item We introduce the \textbf{Focus-Then-Verify} reward architecture, which targets \textbf{sentence-level entailment} within focused context windows to resolve the signal dilution inherent in document-level scoring.
\item We show that EvidenceRL improves accuracy and evidence grounding across model scales, reducing hallucinations and increasing evidence-supported predictions on high-stakes medical and legal domains.
\end{itemize}

\begin{figure*}
    \centering
    \includegraphics[width=0.98\linewidth]{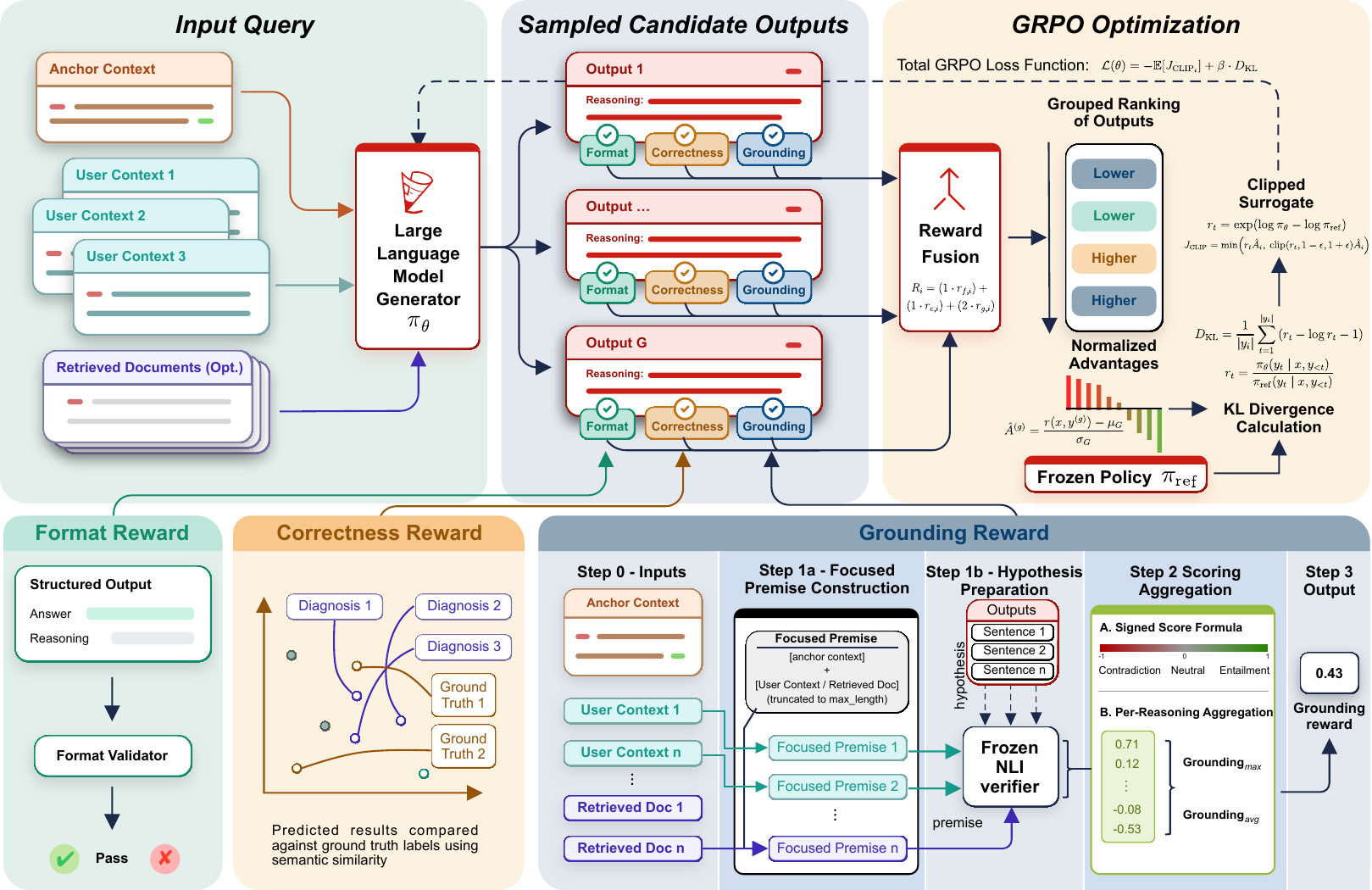}
    \caption{EvidenceRL aligns task accuracy with faithful evidence use across domains. Training uses GRPO with rewards for correctness ($r_c$), format ($r_f$), and evidence grounding ($r_g$). Grounding is computed via a \emph{Focus–Then–Verify} procedure: (1) focused (premise, hypothesis) pairs are constructed by combining an anchor context with individual evidence sections, and (2) each pair is scored by a frozen NLI cross-encoder. 
    }
    \label{fig:framework}
\end{figure*}
\section{Related Work}

\subsection{Evidence-Grounded Generation}
RAG conditions LLM outputs on retrieved documents \cite{lewis2020retrieval}, yet hallucinations remain common even in production systems \cite{huang2025survey}. A central failure mode is \emph{unconstrained generation}: models often produce answers that are correct in isolation but unsupported by the supplied evidence, a phenomenon termed \emph{post-rationalization} \cite{wallat2024correctness}. Citation-based metrics frequently fail to detect this behavior because models can attach superficially related passages without relying on them during reasoning \cite{liu2023evaluating}. Similar issues appear in explanation and chain-of-thought generation, where produced rationales do not reflect the information actually used by the model \cite{turpin2023language}.

Recent work attributes these failures to competition between \emph{parametric knowledge} and \emph{contextual evidence}. When the two conflict, models often default to memorized associations rather than retrieved information \cite{xu-etal-2024-knowledge-conflicts,longpre-etal-2021-entity}. Mechanistic studies link this behavior to interactions between feedforward knowledge circuits and attention-based copying mechanisms, where hallucinations arise when parametric pathways dominate contextual signals \cite{sun2025redeep}. Long-contexts exacerbate the problem, as relevant evidence is often attenuated or ignored \cite{liu2024lost}, and models struggle to reason over conflicting retrieved sources \cite{cattan2025dragged}.

Existing solutions largely operate at inference time. Approaches such as Corrective-RAG \cite{yan2024corrective}, Self-RAG \cite{asai2024self}, CiteGuard \cite{choi2025citeguard}, and RARE \cite{tran2025rare} introduce critique, filtering, or iterative retrieval, while FaithfulRAG \cite{zhang-etal-2025-faithfulrag} explicitly models factual conflicts during reasoning. Other work targets transparency or mechanistic intervention \cite{ye2026seeing,shi2024ircan}. While these methods reduce hallucinations, they treat evidence grounding as a \emph{post-hoc correction problem}. As a result, the underlying generation policy remains weakly constrained by evidence.

\subsection{Verification and Post-hoc Fact Checking}

A parallel line of work focuses on detecting unsupported statements after generation. Early evaluation metrics such as ROUGE fail to capture factual inconsistency, motivating NLI-based verification methods such as FactCC \cite{kryscinski2020evaluating}. Subsequent approaches adopt question-answering or claim-level verification formulations (QAGS, QuestEval, SummaC) to better capture semantic grounding \cite{wang2020asking,scialom2021questeval,laban2022summac}. More recent systems rely on LLM-based judges or self-consistency signals, including SelfCheckGPT, semantic entropy, and rubric-guided evaluation \cite{manakul2023selfcheckgpt,farquhar2024detecting,liu2023g}. Tool-based pipelines such as FacTool and FactScore further decompose generations into atomic facts for verification \cite{chern2023factool,min2023factscore}.

While these methods improve hallucination detection, verification alone does not alter generation behavior.
Models can produce answers from parametric memory and attach superficially supportive evidence afterward \cite{gao2023enabling}. Attribution benchmarks such as ALCE formalize this gap between correctness and groundedness. While recent work begins to optimize against verification signals \cite{tang2025ssfo,xu2025beyond}, most approaches remain evaluation frameworks rather than directly enforcing evidence-consistent generation.

\subsection{RL for Truthfulness and Grounding}

Reinforcement learning has emerged as a scalable mechanism for aligning LLM behavior. RLHF improves helpfulness and safety \cite{ouyang2022training}, but human feedback is poorly calibrated for factual accuracy and uncertainty, often rewarding plausibility over verifiability \cite{augenstein2024factuality,casper2023open}. This misalignment leads to overconfident answers and weak grounding in evidence \cite{xiao2025restoring}. Methods such as TruthRL address this by explicitly rewarding epistemic humility and boundary awareness \cite{wei2025truthrl}.

More recent work replaces human feedback with automated factuality signals. Preference optimization using factuality metrics such as FactScore has been shown to outperform human-labeled rewards \cite{tian2023fine}, while aggregated verifier ensembles improve reward stability \cite{ye2025optimising}. In retrieval-augmented settings, RL has been used to encourage citation use and evidence-grounded answers, beginning with WebGPT and GopherCite \cite{nakano2021webgpt,menick2022teaching} and extending to modern RAG systems.

However, most existing methods optimize \emph{outcome correctness} rather than \emph{evidence consistency}. Context-DPO \cite{bi2025context} improves robustness under conflicting contexts and PA-RAG \cite{wu2025pa} introduces sequential preference optimization for citation quality. 
Despite this progress, reward signals are typically coarse (document-level or answer-level), which allows models to remain correct while relying on unsupported reasoning.
EvidenceRL builds on this line of work but targets a more fundamental objective: enforcing that model outputs remain \emph{consistent with the provided evidence at the level of individual claims}. 

\section{Methodology: EvidenceRL}
\label{sec:method}

\subsection{Setup and Notation}
Let $\mathcal{D}$ be a dataset of pairs $(x, y^\star)$, where $x$ is the input context and $y^\star$ is the reference answer. The generator is an autoregressive policy $\pi_\theta$:
\begin{equation}
  \pi_\theta(y \mid x)
  =
  \prod_{t=1}^{T}
  \pi_\theta(y_t \mid y_{<t}, x).
  \label{eq:policy}
\end{equation}
Each sampled output $y$ contains (i) an explicit reasoning segment $r(y)$ and (ii) final predictions.

\subsection{Reward}
\label{sec:reward}

We decompose the training reward into three components that jointly encourage well-structured, evidence-grounded, and diagnostically accurate outputs.

\paragraph{Format Reward.}
Because downstream grounding verification requires parseable structured reasoning, we include a binary format signal:
\begin{equation}
  r_f(y)
  =
  \begin{cases}
    1 & \text{if } y \text{ is valid JSON },\\
    0 & \text{otherwise.}
  \end{cases}
  \label{eq:format}
\end{equation}

\paragraph{Grounding Reward: Focus--Then--Verify Decomposition.}

Assessing whether reasoning is \emph{grounded} is challenging in multi-source settings, as concatenating all evidence into a single NLI premise causes context overflow and signal dilution. We therefore adopt a \textbf{focus--then--verify} strategy that performs targeted natural language inference (NLI) checks against individual evidence sources. Let $a(x)$ denote an \emph{anchor} context capturing the core framing of the input (chief complaint and history of present illness). For each supplementary section $s_i \in \mathcal{S}(x)$ (e.g., physical exam, imaging reports), we construct a focused premise:
\begin{equation}
\mathcal{P}(x)
=
\{a(x) \oplus s_i\}_{i=1}^{|\mathcal{S}|}.
\end{equation}
When retrieval-augmented generation is used, retrieved documents $\{e_j\}_{j=1}^{k}$ are added as additional premises:
\begin{equation}
\mathcal{P}(x)
\leftarrow
\mathcal{P}(x)
\cup
\{a(x) \oplus e_j\}_{j=1}^{k}.
\end{equation}
For each premise $p \in \mathcal{P}(x)$ we compute a support signal using a frozen NLI model:
\begin{equation}
\begin{aligned}
\Delta_{\text{NLI}}(p, r(y))
&= P(\text{entail}\mid p, r(y)) \\
&\quad - P(\text{contradict}\mid p, r(y)).
\end{aligned}
\label{eq:nli}
\end{equation}

We aggregate over premises by taking the score with the largest magnitude (sign preserved), capturing the strongest supporting or contradicting evidence:
\begin{align}
r^{\max}_g(x,y)
=
\Delta_{\text{NLI}}(p^\star, r(y)), \\
p^\star =  \text{argmax}_{p \in \mathcal{P}(x)} \left|\Delta_{\text{NLI}}(p, r(y))\right|.
\label{eq:grounding_max}
\end{align}
We also report a complementary \emph{average grounding score} measuring overall consistency across sources:
\begin{equation}
r_g^{\text{avg}}(x,y)
=
\frac{1}{|\mathcal{P}(x)|}
\sum_{p \in \mathcal{P}(x)}
\Delta_{\text{NLI}}(p, r(y)).
\label{eq:grounding_avg}
\end{equation}

\paragraph{Correctness Reward.}
During training, correctness is estimated using a lightweight proxy rather than an LLM judge to keep online RL efficient. For open-ended tasks (e.g., clinical diagnosis), predicted answers are embedded with a domain-specific encoder and compared to references via cosine similarity:
\begin{equation}
r_c(x,y) =
\frac{1}{|\hat{\mathcal{Y}}|}
\sum_{\hat{y} \in \hat{\mathcal{Y}}}
\mathbf{1}\!\left[
\max_{y^\star \in \mathcal{Y}^\star}
\cos\!\left(\phi(\hat{y}), \phi(y^\star)\right) > \tau
\right],
\label{eq:correctness}
\end{equation}
where $\hat{\mathcal{Y}}$ denotes the predicted answers, $\mathcal{Y}^\star$ the reference set, $\phi(\cdot)$ the embedding function, and $\tau$ a similarity threshold. For tasks with discrete answer sets (e.g., multiple-choice legal reasoning), correctness reduces to exact matching with the reference.

\paragraph{Combined Reward.}

The final training reward combines all three components with scalar weights:
\begin{equation}
r(x,y)
=
w_f \cdot r_f(y)
\;+\;
w_c \cdot r_c(x,y)
\;+\;
w_g \cdot \tilde{r}_g(x,y),
\label{eq:reward}
\end{equation}
where $\tilde{r}_g = (r_g + 1)/2$ is the normalized grounding score, and we set $w_f = w_c = 1$, $w_g = 2$ to emphasize evidence grounding, our primary objective. 

\subsection{Optimization with GRPO}
We optimize $\pi_\theta$ using Group Relative Policy Optimization (GRPO)~\citep{shao2024deepseekmath}. For each input $x$, we sample a group of $G$ completions from the current policy, $\{y^{(1)}, \ldots, y^{(G)}\} \sim \pi_{\theta_{\text{old}}}(\cdot \mid x)$.
Each completion is scored using the reward in Eq.~\eqref{eq:reward}. Rewards are normalized within the group to obtain advantages:
\begin{equation}
\hat{A}^{(g)} =
\frac{r(x,y^{(g)}) - \mu_G}{\sigma_G},
\label{eq:advantage}
\end{equation}
where $\mu_G$ and $\sigma_G$ are the mean and standard deviation across the group. 
Following standard GRPO, the policy is updated using a clipped policy-gradient objective with KL regularization against a frozen reference policy $\pi_{\text{ref}}$, with advantages applied uniformly across tokens of each sampled completion.

\section{Experiments}
\label{sec:experiments}

\subsection{Task and Dataset}

We evaluate EvidenceRL in two high-stakes domains, medicine and law. Medical diagnosis requires synthesizing clinical findings into ranked hypotheses, whereas legal analysis involves applying statutory rules to case facts.

\paragraph{MIMIC-IV-Ext (Cardiac).}
We use de-identified ICU cases from MIMIC-IV-Ext \citep{johnson2023mimic}. Ground-truth labels are derived from ICD-10 cardiac codes (prefix “I”). Given patient context $x$ and optional retrieved evidence $E_{\text{pre}}$, the model predicts five ranked diagnoses with supporting reasoning. We split the dataset into 3{,}700 training and 1{,}000 held-out cases (Appendix~\ref{app:dataset}).

\paragraph{BarExam MBE (Legal).}
We use the Multistate Bar Examination benchmark \citep{zheng2025reasoning}, where each instance contains a fact pattern, four answer choices, and a gold legal passage as evidence. Given context $x$ and retrieved authority $E_{\text{pre}}$, the model selects the correct answer with grounded reasoning. The dataset includes 954 training and 173 test cases across six legal subjects.

\paragraph{Knowledge Source.}
For both tasks, models can receive retrieved domain-specific evidence $E_{\text{pre}}$ to support grounded reasoning (Appendix~\ref{app:retrieval}).

\subsection{Models and Baselines}
We evaluate eight backbones: Gemma-3 (4B, 12B, 27B), Llama (3.2-3B, 3.1-8B, 3.3-70B), and GPT-oss (20B, 120B). All methods use the same structured reasoning prompt (Appendix~\ref{app:prompts}).

We compare six approaches. Reasoning Only performs direct diagnosis generation from patient context without retrieval. Self-RAG performs adaptive retrieval and evidence critique during generation~\citep{asai2024self}. Self-Consistency samples $N$ completions and aggregates via semantic clustering and majority voting~\citep{wang2022self}.
SFT applies supervised fine-tuning on curated diagnosis–reasoning pairs (Appendix~\ref{app:models}).
Faithful DPO (fDPO) constructs cross-model preference pairs based purely on NLI grounding quality, without any correctness signal, in the spirit of context-faithful preference optimization~\citep{bi2025context}. 
EvidenceRL applies GRPO training with the combined reward in Eq.~\ref{eq:reward}, jointly optimizing both grounding and correctness. All fine-tuning uses LoRA~\citep{hu2022lora} (Appendix~\ref{app:training_appendix}).

\subsection{Metrics}
\label{sec:metrics}


\paragraph{Task Correctness.}
We report F1@$k$ or Accuracy. For MIMIC, correctness is determined by an LLM judge evaluating semantic equivalence between predicted and ground-truth diagnoses, $\text{Judge}(\hat{y}_i, y^\star_j)$.

\paragraph{Evidence Grounding.}
We report Grounding@$k$ using the NLI scores $r_g^{\text{max}}$ and $r_g^{\text{avg}}$ (Eqs.~\ref{eq:grounding_max}–\ref{eq:grounding_avg}), averaged over the top-$k$ predictions.

\paragraph{Diagnostic Taxonomy.}
We classify each prediction into a $3 \times 2$ taxonomy based on correctness and grounding strength ($r_g^{\max}$). Grounded ($r_g^{\max} > 0.5$), Weak ($r_g^{\max} \in [-0.5, 0.5]$) and Contradicted ($r_g^{\max} < -0.5$):
\begin{center}
\small
\resizebox{\linewidth}{!}{
\begin{tabular}{l|c|c}
\toprule
& \textbf{Correct Answer} & \textbf{Incorrect Answer} \\
\midrule

\textbf{Grounded}
  & \cellcolor{goodgreen} Evidence-Based
  & \cellcolor{warnyellow} Grounded Error \\

\textbf{Weak}
  & \cellcolor{neutralyellow} Weakly Supported
  & \cellcolor{neutralyellow} Unsupported Error \\

\textbf{Contradicted}
  & \cellcolor{lightorange} Lucky Guess
  & \cellcolor{badred} Hallucination \\

\bottomrule
\end{tabular}
}
\end{center}






We report key rates: EB\%, H\%, LG\%, and Weak\%. We additionally compute faithfulness, which measures how often correct predictions are genuinely evidence-supported.
\begin{equation}
\text{Faithfulness}
=
\frac{\text{EB}}{\text{EB} + \text{WS} + \text{LG}},
\label{eq:faithful}
\end{equation}
\begin{table*}[!t]
\centering
\caption{
Diagnostic performance and evidential reliability ($\pm$95\% bootstrap CI). We report F1@3, average grounding ($G_{\mathrm{avg}}@3$), and a diagnostic taxonomy decomposing predictions into evidence-based correctness (EB), hallucinations (H), and lucky guesses (LG). Faithfulness (F) measures the fraction of correct predictions supported by evidence (Eq.~\ref{eq:faithful}). Green marks the best backbone within each method; yellow marks the overall best.
}
\label{tab:main_results}

\setlength{\tabcolsep}{6.5pt}
\renewcommand{\arraystretch}{0.78}

\resizebox{\textwidth}{!}{%
\begin{tabular}{c |l| c| c c | c c c c | c}
\toprule
\rowcolor{headerblue}
\textbf{Method} &
\textbf{Backbone} &
\textbf{F1@3} ($\uparrow$) &
\textbf{$G_{\mathrm{avg}}$@3} ($\uparrow$) &
\textbf{$G_{\mathrm{max}}$@3} ($\uparrow$) &
\textbf{EB} ($\uparrow$) &
\textbf{H} ($\downarrow$)&
\textbf{W} ($\downarrow$)&
\textbf{LG} ($\downarrow$)&
\textbf{F} ($\uparrow$)\\
\midrule

\multirow{8}{*}{\textbf{\shortstack{Reasoning \\ Only}}}
& Llama-3.2-3B  & 37.0 $\pm$ 1.6 & \best{45.3 $\pm$ 3.5} & 47.6 $\pm$ 3.6 & 31.8\% & 11.6\% & 16.5\% & \best{6.2\%} & \best{72.5\%} \\
& Llama-3.1-8B  & 38.2 $\pm$ 1.3 & 19.4 $\pm$ 3.6 & 21.3 $\pm$ 3.7 & 27.4\% & 17.7\% & 20.2\% & 11.5\% & 56.6\% \\
& Llama-3.3-70B & 51.3 $\pm$ 1.3 & \best{46.9 $\pm$ 3.0} & \best{49.4 $\pm$ 3.1} & \best{43.8\%} & \best{8.5\%} & \best{14.3\%} & 9.3\% & 71.9\% \\
& Gemma-3-4B    & 44.2 $\pm$ 1.4 & 34.3 $\pm$ 3.3 & 36.3 $\pm$ 3.4 & 35.1\% & 13.4\% & 19.2\% & 8.9\% & 66.0\% \\
& Gemma-3-12B   & 46.5 $\pm$ 1.3 & 21.5 $\pm$ 3.2 & 23.5 $\pm$ 3.4 & 31.0\% & 14.4\% & 19.2\% & 14.3\% & 56.3\% \\
& Gemma-3-27B   & \best{52.2 $\pm$ 1.3} & 37.1 $\pm$ 3.2 & 39.0 $\pm$ 3.4 & 39.1\% & \best{8.5\%} & 15.5\% & 13.7\% & 63.1\% \\
& GPT-oss-20B   & 46.0 $\pm$ 1.3 & -2.2 $\pm$ 3.4 & -0.9 $\pm$ 3.6 & 24.9\% & 15.6\% & 17.7\% & 26.0\% & 40.2\% \\
& GPT-oss-120B  & 43.6 $\pm$ 1.3 & 18.3 $\pm$ 3.4 & 19.7 $\pm$ 3.5 & 30.8\% & 14.4\% & 15.5\% & 17.9\% & 53.3\% \\
\midrule

\multirow{8}{*}{\textbf{Self-RAG}}
& Llama-3.2-3B  & 36.5 $\pm$ 1.5 & 24.4 $\pm$ 4.4 & 25.6 $\pm$ 4.6 & 26.0\% & 17.3\% & 18.5\% & 10.4\% & 59.6\% \\
& Llama-3.1-8B  & 39.4 $\pm$ 1.3 & 30.0 $\pm$ 3.3 & 33.1 $\pm$ 3.5 & 31.5\% & 13.6\% & 19.9\% & 9.6\% & 62.7\% \\
& Llama-3.3-70B & 51.6 $\pm$ 1.3 & 39.9 $\pm$ 3.1 & 42.1 $\pm$ 3.3 & \best{41.0\%} & 9.0\% & \best{16.3\%} & 11.6\% & 66.3\% \\
& Gemma-3-4B    & 46.5 $\pm$ 1.3 & \best{46.2 $\pm$ 3.1} & \best{48.7 $\pm$ 3.2} & 39.2\% & \best{8.2\%} & 18.1\% & \best{7.8\%} & \best{70.4\%} \\
& Gemma-3-12B   & 47.0 $\pm$ 1.3 & 17.5 $\pm$ 3.2 & 19.0 $\pm$ 3.4 & 29.5\% & 14.4\% & 19.3\% & 16.2\% & 52.9\% \\
& Gemma-3-27B   & \best{53.3 $\pm$ 1.3} & 25.8 $\pm$ 3.4 & 27.4 $\pm$ 3.5 & 34.5\% & 9.5\% & 17.5\% & 17.7\% & 54.3\% \\
& GPT-oss-20B   & 36.5 $\pm$ 1.6 & 0.2 $\pm$ 3.0 & 1.7 $\pm$ 3.1 & 25.7\% & 16.3\% & 17.6\% & 23.7\% & 42.7\% \\
& GPT-oss-120B  & 37.1 $\pm$ 1.6 & 11.4 $\pm$ 3.1 & 12.9 $\pm$ 3.2 & 29.9\% & 15.7\% & 16.4\% & 18.2\% & 52.6\% \\
\midrule

\multirow{8}{*}{\textbf{\shortstack{Self \\ Consistency}}}
& Llama-3.2-3B  & 39.3 $\pm$ 2.2 & 48.2 $\pm$ 4.7 & 50.3 $\pm$ 4.8 & 33.6\% & 10.7\% & 15.6\% & \best{5.9\%} & 71.8\% \\
& Llama-3.1-8B  & 34.8 $\pm$ 1.3 & 25.8 $\pm$ 3.1 & 27.6 $\pm$ 3.3 & 26.8\% & 17.6\% & 19.7\% & 8.7\% & 60.7\% \\
& Llama-3.3-70B & 46.8 $\pm$ 1.3 & \best{51.2 $\pm$ 3.0} & \best{53.2 $\pm$ 3.1} & \best{41.2\%} & 9.2\% & \best{12.1\%} & 7.8\% & 74.5\% \\
& Gemma-3-4B    & 41.9 $\pm$ 1.4 & 48.3 $\pm$ 3.1 & 50.5 $\pm$ 3.2 & 37.3\% & 10.6\% & 14.9\% & 6.2\% & \best{75.3\%} \\
& Gemma-3-12B   & 41.0 $\pm$ 1.3 & 24.4 $\pm$ 3.3 & 26.0 $\pm$ 3.5 & 28.7\% & 16.4\% & 17.0\% & 12.0\% & 59.7\% \\
& Gemma-3-27B   & \best{48.6 $\pm$ 1.3} & 45.9 $\pm$ 3.1 & 48.1 $\pm$ 3.2 & 39.6\% & \best{8.7\%} & 13.9\% & 10.1\% & 68.7\% \\
& GPT-oss-20B   & 44.0 $\pm$ 1.3 & 0.4 $\pm$ 3.5 & 1.6 $\pm$ 3.6 & 26.9\% & 16.5\% & 15.7\% & 24.6\% & 43.5\% \\
& GPT-oss-120B  & 41.2 $\pm$ 1.3 & 25.1 $\pm$ 3.3 & 26.7 $\pm$ 3.4 & 32.3\% & 15.2\% & 16.1\% & 13.0\% & 59.2\% \\
\midrule

\multirow{7}{*}{\textbf{SFT}}
& Llama-3.2-3B  & \best{49.1 $\pm$ 1.3} & \best{34.8 $\pm$ 3.2} & \best{36.5 $\pm$ 3.4} & \best{41.3\%} & \best{11.5\%} & \best{15.5\%} & \best{12.2\%} & \best{67.0\%} \\
& Llama-3.1-8B  & 47.2 $\pm$ 1.4 & 24.9 $\pm$ 3.3 & 26.5 $\pm$ 3.5 & 37.9\% & 13.3\% & 17.5\% & 13.8\% & 61.4\% \\
& Llama-3.3-70B & 35.2 $\pm$ 1.8 & 8.9 $\pm$ 3.0 & 10.0 $\pm$ 3.2 & 32.5\% & 14.3\% & 18.1\% & 19.5\% & 52.1\% \\
& Gemma-3-4B    & 39.9 $\pm$ 1.6 & 2.1 $\pm$ 3.3 & 1.8 $\pm$ 3.4 & 28.4\% & 18.2\% & 18.6\% & 21.1\% & 47.3\% \\
& Gemma-3-12B   & 33.8 $\pm$ 1.8 & 0.3 $\pm$ 3.1 & 0.6 $\pm$ 3.3 & 28.5\% & 16.9\% & 18.8\% & 23.4\% & 45.3\% \\
& Gemma-3-27B   & 36.6 $\pm$ 1.7 & 1.2 $\pm$ 3.1 & 1.2 $\pm$ 3.3 & 27.8\% & 18.0\% & 19.7\% & 21.5\% & 45.7\% \\
& GPT-oss-20B   & 47.7 $\pm$ 1.4 & 31.8 $\pm$ 3.4 & 33.1 $\pm$ 3.6 & 38.0\% & 12.1\% & 15.7\% & 12.8\% & 63.5\% \\
\midrule

\multirow{6}{*}{\textbf{fDPO}}
& Llama-3.2-3B  & 38.7 $\pm$ 1.4 & 62.4 $\pm$ 2.7 & 63.6 $\pm$ 2.8 & 37.9\% & 7.9\% & 11.4\% & 3.6\% & 82.2\% \\
& Llama-3.1-8B  & 39.7 $\pm$ 1.4 & 51.4 $\pm$ 3.4 & 53.0 $\pm$ 3.5  & 39.5\% & 9.0\% & 11.2\% & 7.3\% & 76.7\% \\
& Gemma-3-4B    & 46.3 $\pm$ 1.3 & \best{68.9 $\pm$ 2.6} & \best{70.8 $\pm$ 2.6} & \best{47.6\%} & \best{5.9\%} & \best{9.8\%} & \best{3.1\%} & \best{86.4\%} \\
& Gemma-3-12B   & 47.9 $\pm$ 1.3 & 37.6 $\pm$ 3.1 & 39.9 $\pm$ 3.3 & 37.0\% & 10.2\% & 16.5\% & 11.0\% & 65.6\% \\
& Gemma-3-27B   & \best{53.1 $\pm$ 1.3} & 62.5 $\pm$ 2.7 & 64.6 $\pm$ 2.8 & 48.5\% & 4.4\% & 11.6\% & 7.0\% & 77.7\% \\
& GPT-oss-20B   & 46.0 $\pm$ 1.3 & 14.5 $\pm$ 3.4 & 15.7 $\pm$ 3.6 & 29.9\% & 13.1\% & 16.5\% & 20.8\% & 48.4\% \\
\midrule

\multirow{5}{*}{\textbf{\shortstack{EvidenceRL \\ with GRPO \\ (Ours)}}}
& Llama-3.2-3B  & \best{54.5} $\pm$ 1.3 & \overallbest{77.0 $\pm$ 2.2} & \overallbest{78.7 $\pm$ 2.2} & \overallbest{61.6\%} & \overallbest{2.4\%} & \overallbest{8.6\%} & \overallbest{3.4\%} & \overallbest{87.5\%} \\
& Llama-3.1-8B  & 53.9 $\pm$ 1.4 & \best{58.9 $\pm$ 2.9} & \best{61.1 $\pm$ 3.0} & \best{55.3\%} & \best{4.3\%} & 14.2\% & 7.2\% & \best{76.1\%} \\
& Gemma-3-4B    & 46.8 $\pm$ 1.4 & 54.0 $\pm$ 3.0 & 56.0 $\pm$ 3.1 & 42.4\% & 7.8\% & \best{13.9\%} & \best{6.6\%} & 76.4\% \\
& Gemma-3-12B   & 49.3 $\pm$ 1.3 & 38.0 $\pm$ 3.1 & 40.5 $\pm$ 3.3 & 38.0\% & 10.6\% & 17.2\% & 10.3\% & 65.9\% \\
& Gemma-3-27B   & \overallbest{54.9 $\pm$ 1.3} & 46.2 $\pm$ 3.1 & 48.4 $\pm$ 3.2 & 44.4\% & 7.0\% & 14.1\% & 11.4\% & 69.0\% \\
\bottomrule
\end{tabular}
}
\end{table*}
\section{Results}
\label{sec:results}

We evaluate EvidenceRL across two structurally distinct high-stakes domains, clinical diagnosis and legal reasoning, where ungrounded predictions carry real-world consequences, measuring both diagnostic accuracy and evidential reliability. 

\begin{table*}[!t]
\centering
\caption{BarExam MBE: Accuracy, Evidence Grounding, and Diagnostic Taxonomy across backbone models.
Wider confidence intervals reflect the smaller test($n{=}173$) set compared to the medical domain ($n{=}1{,}000$).
}
\label{tab:barexam_results}
\setlength{\tabcolsep}{6.5pt}
\renewcommand{\arraystretch}{0.78}

\resizebox{\textwidth}{!}{%
\begin{tabular}{l|l|c|cc|cccccc}
\toprule
\rowcolor{headerblue} \textbf{Method} &
\textbf{Backbone} & \textbf{Acc.}($\uparrow$) & \textbf{$G_{\text{avg}}$}($\uparrow$) & \textbf{$G_{\max}$}($\uparrow$) & \textbf{EB}($\uparrow$) & \textbf{H}($\downarrow$) & \textbf{W}($\downarrow$) & \textbf{LG}($\downarrow$) & \textbf{F}($\uparrow$) \\
\midrule
\multirow{8}{*}{\textbf{\shortstack{Reasoning \\ Only}}}
& Llama-3.2-3B  & 42.7 $\pm$ 9.0 & -3.8 $\pm$ 7.2  &  5.3 $\pm$ 13.6 & 13.7\% & 15.4\% & 46.2\% & 9.4\%  & 32.0\% \\
& Llama-3.1-8B  & 57.3 $\pm$ 9.4 & -1.7 $\pm$ 8.6  &  3.3 $\pm$ 12.2 & 18.8\% & 10.3\% & 51.3\% & 12.8\% & 32.8\% \\
& Llama-3.3-70B & \overallbest{82.1 $\pm$ 6.8} & \best{13.9 $\pm$ 4.8}  & \best{49.5 $\pm$ 12.4} & \overallbest{50.4\%} & \overallbest{1.7\%}  & \best{30.8\%} & \best{6.0\%}  & \best{61.5\%} \\
& Gemma-3-4B    & 47.9 $\pm$ 8.6 &  5.3 $\pm$ 6.6  & 14.5 $\pm$ 14.8 & 19.7\% & 10.3\% & 44.4\% & 6.8\%  & 41.1\% \\
& Gemma-3-12B   & 60.7 $\pm$ 8.5 &  4.2 $\pm$ 4.6  & 21.0 $\pm$ 15.3 & 22.2\% & 5.1\%  & 43.6\% & 12.0\% & 36.6\% \\
& Gemma-3-27B   & 65.8 $\pm$ 8.5 &  4.5 $\pm$ 4.4  & 20.2 $\pm$ 15.7 & 28.2\% & 7.7\%  & 45.3\% & 7.7\%  & 42.9\% \\
& GPT-oss-20B   & 54.7 $\pm$ 8.5 & -8.3 $\pm$ 8.1  & -1.5 $\pm$ 14.7 & 15.4\% & 12.8\% & 53.8\% & 9.4\%  & 28.1\% \\
& GPT-oss-120B  & 77.8 $\pm$ 7.7 &  1.0 $\pm$ 6.0  &  7.7 $\pm$ 15.8 & 33.3\% & 9.4\%  & 31.6\% & 17.9\% & 42.9\% \\
\midrule
\multirow{4}{*}{\textbf{\shortstack{EvidenceRL \\ with GRPO \\ (Ours)}}}
& Llama-3.2-3B & 50.4 $\pm$ 9.0 &  7.7 $\pm$ 5.3 & 29.3 $\pm$ 14.4 & 26.5\% & 10.3\% & 36.8\% & 6.0\%  & 55.9\% \\
& Llama-3.1-8B & \best{60.7 $\pm$ 9.0} & \overallbest{18.9 $\pm$ 6.2} & \overallbest{51.9 $\pm$ 12.5} & \best{41.0\%} & \best{6.0\%}  & 26.5\% & \overallbest{4.3\%}  & \overallbest{67.6\%} \\
& Gemma-3-4B   & 48.7 $\pm$ 8.5 &  7.8 $\pm$ 4.4 & 33.8 $\pm$ 15.2 & 26.5\% & \best{6.0\%}  & 39.3\% & 6.8\%  & 54.4\% \\
& Gemma-3-12B  & \best{60.7 $\pm$ 8.5} & 11.0 $\pm$ 5.9 & 33.6 $\pm$ 15.2 & 37.6\% & 6.8\%  & \overallbest{23.1\%} & 9.4\%  & 62.0\% \\

\bottomrule
\end{tabular}

}
\end{table*}

\subsection{EvidenceRL Reduces the Faithfulness Gap and Narrows the Scale Gap}

A persistent challenge in high-stakes domains is the \emph{faithfulness gap}: models achieve high task accuracy through parametric recall rather than evidence-grounded reasoning. This pattern appears in both domains. In medicine (Table~\ref{tab:main_results}), Llama-3.1-8B achieves 56.6\% F1@3 but only 19.4 $G_{\mathrm{avg}}@3$, indicating that many correct diagnoses are unsupported by patient evidence. The legal domain (Table~\ref{tab:barexam_results}) shows an even stronger version of this effect: under reasoning-only inference, grounding is negative for several models (e.g., $G_{\mathrm{avg}}=-3.8$ for Llama-3.2-3B and $-8.3$ for GPT-oss-20B). Even the strongest baseline, Llama-3.3-70B (Acc 82.1\%, F 61.5\%), exhibits only moderate grounding ($G_{\mathrm{avg}}=13.9$) with an EB rate of 50.4\%, meaning nearly half of correct predictions remain non-evidential.

EvidenceRL training fundamentally shifts the \emph{basis} of correct predictions. In the medical domain, GRPO increases Llama-3.2-3B’s EB correctness from 31.8\% to 61.6\% while reducing hallucinations and lucky guesses to 2.4\% and 3.4\%, and raising F1@3 from 37.0 to 54.5. A similar redistribution appears in the legal domain: Llama-3.1-8B’s EB rate rises from 18.8\% to 41.0\%, hallucinations fall from 10.3\% to 6.0\%, lucky guesses from 12.8\% to 4.3\%, and faithfulness increases from 32.8\% to 67.6\% alongside $G_{\mathrm{avg}}$ improving from $-1.7$ to $18.9$. Across both domains, EvidenceRL provides a training signal that penalizes unsupported correctness, moving correct predictions into the explicitly evidence-grounded category.

This shift also narrows the traditional scaling advantage of larger models. Under zero-shot reasoning, larger models are not consistently more evidence-grounded (e.g., Gemma-3-4B achieves higher grounding than Gemma-3-12B in both domains). After EvidenceRL training, smaller models can surpass much larger baselines while remaining more grounded: in medicine, EvidenceRL Llama-3.2-3B (F1@3 = 54.5) exceeds reasoning-only Llama-3.3-70B (51.3) and Gemma-3-27B (52.2); in law, EvidenceRL Llama-3.2-3B (F = 55.9\%) surpasses several larger reasoning-only models. These results indicate that improving \emph{how} models use evidence during training can partially substitute for parameter scale, offering a compute-efficient path to reliable reasoning in high-stakes settings.

\subsection{When the Objective Is Not Aligned: SFT and Inference-Time Controls}




Methods that do not modify the learning objective fail to reliably align diagnostic accuracy with evidence grounding. Supervised Fine-Tuning (SFT) preserves plausible predictions but collapses grounding: across the Gemma family, $G_{\mathrm{avg}}@3$ falls nearly to zero (e.g., 37.1 $\rightarrow$ 1.2 on Gemma-3-27B) while F1@3 remains similar (36.6). This indicates that SFT teaches models to imitate answer format and cite evidence without learning the semantic relationship between diagnoses and supporting text.

Inference-time controls show a different but similarly limited effect. Techniques such as Self-RAG and Self-Consistency occasionally improve grounding but fail to consistently improve both accuracy and evidence use. For example, on Llama-3.2-3B, Self-RAG slightly reduces F1@3 (37.0 $\rightarrow$ 36.5) while substantially lowering grounding (45.3 $\rightarrow$ 24.4), while Self-Consistency improves grounding through aggregation but retains a higher fraction of reasoning failures. These patterns suggest that inference-time methods primarily redistribute outputs within the model’s existing preference structure, whereas EvidenceRL alters the learning objective itself through reward-based credit assignment, enabling simultaneous gains in diagnostic accuracy and evidential reliability.

\subsection{Comparing Alignment Objectives: fDPO vs.\ EvidenceRL}


Both fDPO and EvidenceRL improve evidence grounding over SFT and inference-time baselines, but they optimize the accuracy–grounding trade-off through different mechanisms.

fDPO trains on cross-model preference pairs, where chosen responses come from the most grounded model outputs and rejected responses from the least grounded. This provides a strong offline grounding signal (mean grounding +0.75 vs.\ −0.55), but also introduces implicit cross-model distillation: weaker backbones may be trained to imitate responses generated by stronger models. On Llama-3.2-3B (Table \ref{tab:main_results}) this results in high Faithfulness (82.2\%) but little accuracy improvement (F1@3 = 38.7 vs.\ 37.0 baseline). The Evidence-Based rate increases only modestly (31.8\%→37.9\%), while Weak and Lucky Guess predictions shrink (16.5\%→11.4\%, 6.2\%→3.6\%). Faithfulness rises primarily because fewer predictions remain correct rather than because more grounded correct predictions are produced.

EvidenceRL avoids this confound by computing rewards on-policy. GRPO evaluates grounding and correctness directly on the model’s own rollouts, rewarding improvements relative to its own candidate generations. As a result, grounded correctness expands rather than contracts. On Llama-3.2-3B, the Evidence-Based rate nearly doubles (31.8\%→61.6\%) while diagnostic accuracy also increases (F1@3 = 54.5), yielding the highest Faithfulness (87.5\%). Here, improvements arise from generating more predictions that are both correct and evidence-supported.

\subsection{Reward Proxy Alignment Analysis}
\begin{figure}[h]
\centering
\includegraphics[width=\columnwidth]{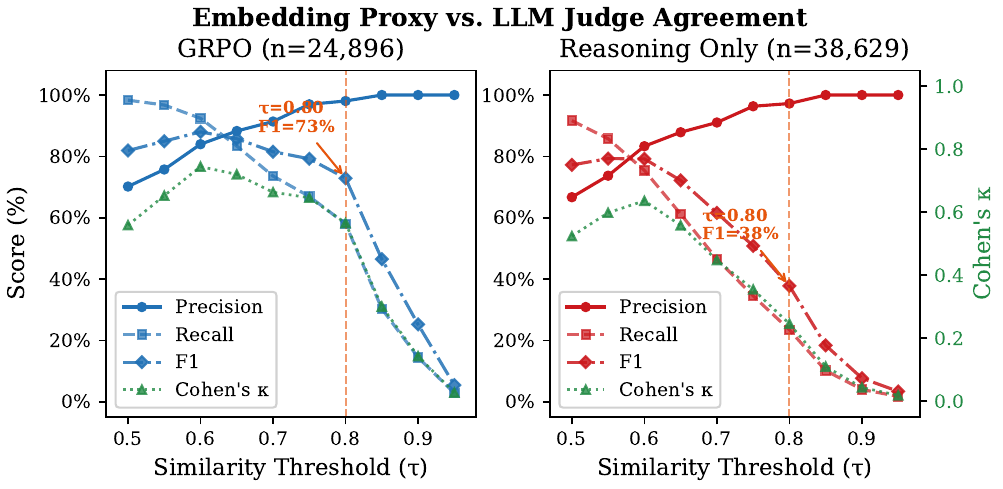}
\caption{
At $\tau=0.80$, precision remains high for both approaches. Stable recall and high Cohen’s $\kappa$ indicate a conservative reward signal, with no evidence of proxy hacking by GRPO models.
}
\label{fig:reward-proxy-alignment}
\end{figure}

Figure~\ref{fig:reward-proxy-alignment} compares the BioLORD-2023 similarity threshold ($\tau=0.80$) against a judge's binary verdicts. Across 24,896 pairs from our GRPO-trained models, the embedding proxy achieved 98.0\% precision, preventing the RL policy from being falsely rewarded for incorrect predictions. The rare false positives (145 cases, or 0.58\%) were legitimate clinical near-misses (e.g., "Left Bundle Branch Block" vs. "Atrioventricular and left bundle-branch block") rather than systematic exploitation. Furthermore, untrained reasoning baselines demonstrated a nearly identical precision (97.2\% across 38,629 pairs), confirming that the optimization process does not learn to hack the proxy metric. Although the conservative $\tau=0.80$ threshold limits recall to 58.0\% by under-crediting non-literal synonyms, this asymmetry safely prioritizes strict error penalization over rewarding diverse phrasing.


\begin{figure}[!h]
\centering
\includegraphics[width=\columnwidth]{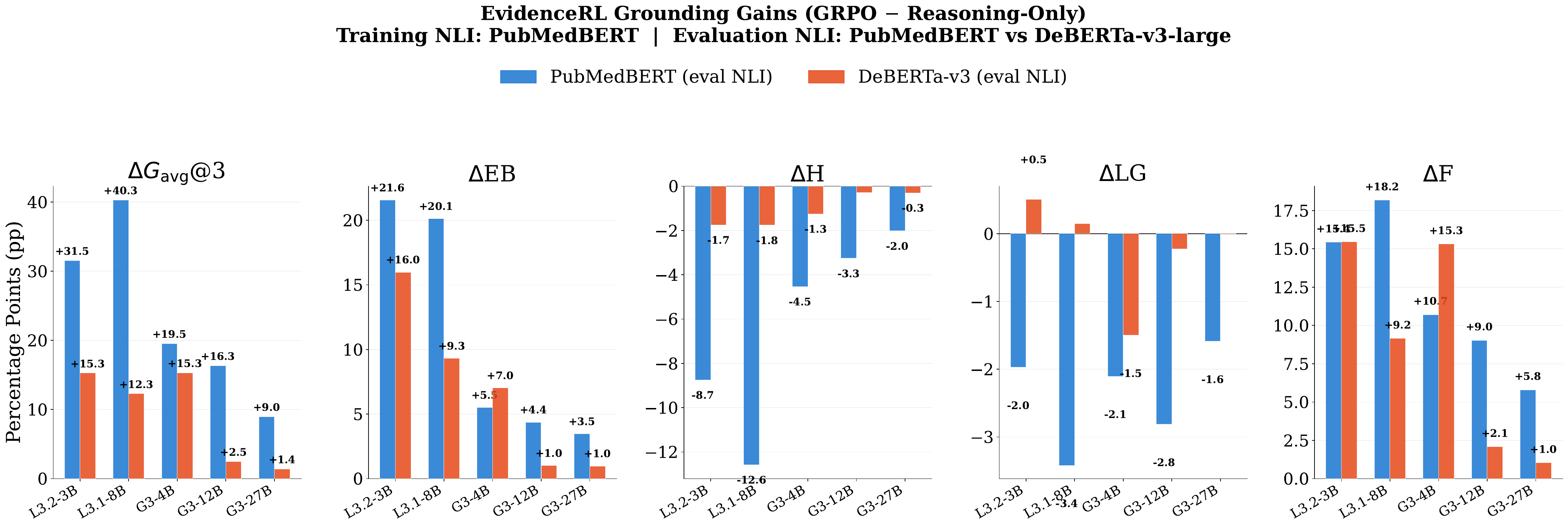}
\caption{All five metrics shift in the same direction under both evaluators, confirming that grounding improvements reflect genuine evidence use rather than reward model overfitting.
}
\label{fig:nli_robustness}
\end{figure}

To ensure grounding improvements reflect genuine evidence use rather than overfitting to the training NLI reward model, we re-evaluated all models using an independent evaluator (DeBERTa-v3-large). As shown in Figure~\ref{fig:nli_robustness}, grounding gains persist across all five backbones: Evidence-Based predictions increased by 1.0–16.0 pp, Hallucinations decreased by 0.3–1.8 pp, and Faithfulness improved by 1.0–15.5 pp. Although absolute magnitudes are smaller, likely due to DeBERTa-v3-large lacking biomedical pretraining, the consistent direction of improvement confirms that GRPO enhances true evidence-grounded reasoning rather than exploiting reward model artifacts.

\subsection{Ablation Study}

\begin{figure}[h]
\centering
\includegraphics[width=\columnwidth]{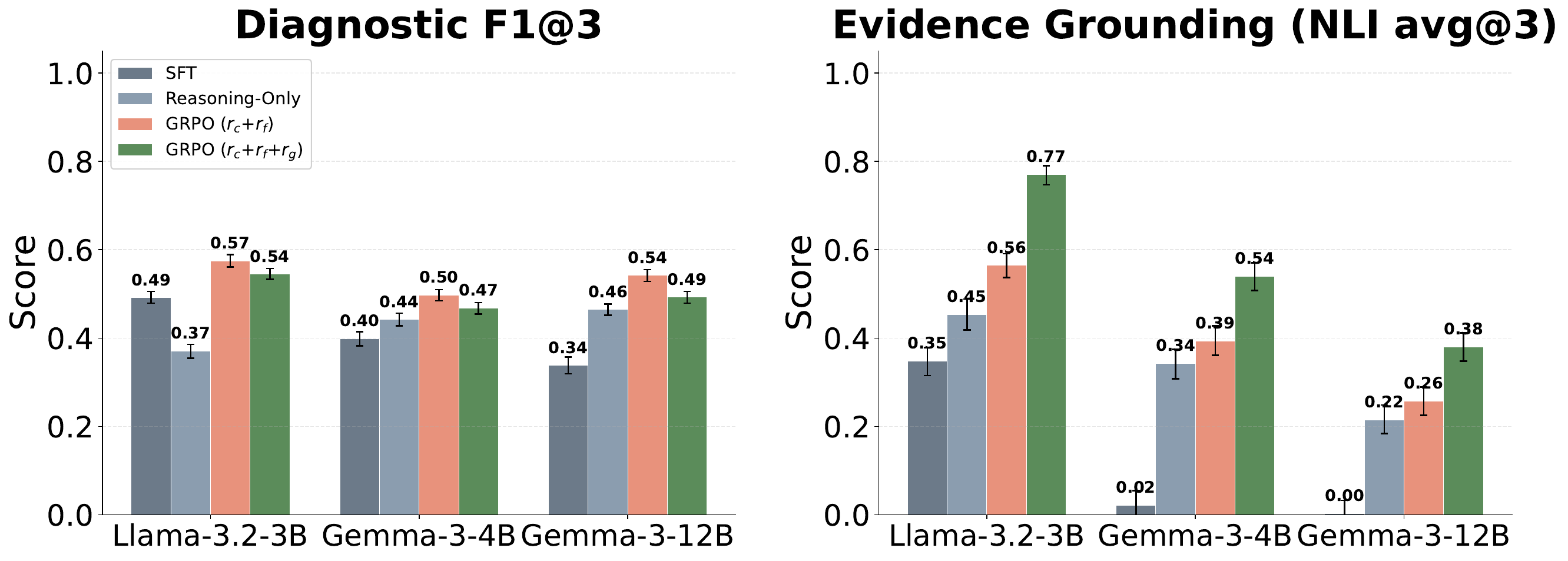}
\caption{
SFT yields reasonable accuracy but weak grounding. GRPO with correctness reward ($r_c{+}r_f$) maximizes F1, while adding the grounding reward ($r_g$) substantially improves evidence attribution with only minor accuracy trade-offs.
}
\label{fig:ablation_main_bars}
\end{figure}

Figure~\ref{fig:ablation_main_bars} isolates the contribution of each reward component across three backbones on the medical domain. SFT produces plausible but ungrounded outputs: while diagnostic precision remains competitive (e.g., P@3=0.614 on Llama-3.2-3B), grounding collapses on the Gemma models (G$*{\text{avg}}$@3 = 0.021 for Gemma-3-4B and 0.003 for Gemma-3-12B), indicating that demonstration tuning teaches citation format without enforcing semantic evidence use. 

RL with correctness reward ($r_c{+}r_f$) yields the highest diagnostic accuracy across all models (F1@3: 0.575, 0.497, 0.542) while moderately improving grounding (0.564, 0.394, 0.258). Adding the grounding reward ($r_g$) substantially increases evidence grounding (0.770, 0.540, 0.380) with only a small reduction in diagnostic accuracy (e.g., 0.575→0.545 on Llama-3.2-3B). Overall, the full reward produces the best accuracy–grounding trade-off, demonstrating that explicit grounding supervision is necessary to align diagnoses with supporting evidence.

\section{Conclusion}
\label{sec:conclusion}

Large language models can produce correct answers while ignoring the evidence provided in context. In high-stakes domains such as clinical diagnosis and legal reasoning, this leads to plausible but unsupported outputs that undermine trust.
EvidenceRL addresses this by making evidence adherence a training objective. Across multiple model families and two domains, it consistently shifts model behavior toward predictions that are both accurate and evidence-grounded while reducing hallucinations and lucky guesses, without sacrificing task performance.
These results highlight a fundamental limitation of inference-time controls: retrieval, self-consistency, and post-hoc verification can filter outputs but cannot change how models reason. Grounding-aware reinforcement learning does, reshaping the model’s decision policy to prefer evidence-supported reasoning.

\section{Limitation}
EvidenceRL relies on automated proxies to scale training and evaluation. Grounding is assessed using a frozen biomedical NLI cross-encoder, which provides a domain-appropriate signal but remains an approximation of true evidential reasoning. We mitigate this through sentence-level verification and by evaluating all methods with the same grounding model, ensuring any systematic bias affects approaches equally. Training cost is higher than SFT because GRPO requires sampling and scoring multiple completions per iteration; inference cost remains unchanged. As with retrieval-augmented systems generally, the method assumes retrieved evidence is reliable, so errors in the knowledge source may propagate into otherwise well-grounded outputs. The fDPO baseline uses cross-model preference pairs, introducing a mild knowledge distillation component alongside the grounding signal; within-model grounding-only optimization would provide a cleaner ablation and is a natural extension of this work. Finally, while we demonstrate consistent improvements across two high-stakes domains, cardiac diagnosis and legal reasoning, evaluating EvidenceRL across additional domains and evidence sources remains an important direction for future study.

\section{Potential risks}
EvidenceRL is a research contribution toward more trustworthy AI reasoning and is not intended for autonomous deployment in clinical or legal settings. Responsible use in these domains would require prospective validation, regulatory evaluation, and human-in-the-loop oversight.
While EvidenceRL improves evidence grounding, well-cited reasoning may appear authoritative even when the underlying conclusion is incorrect. The framework therefore aims to make model reasoning \emph{verifiable} rather than automatically trustworthy. In practical deployments, evidence citations should support human review rather than replace it.
The grounding reward enforces consistency with retrieved evidence but does not verify the correctness of that evidence itself. Errors or outdated information in the knowledge source may therefore propagate into otherwise well-grounded outputs, making evidence curation and source quality essential.
Finally, training at the scales explored here requires substantial computational resources. Ensuring that evidence-grounded AI tools remain accessible across institutions will require continued work on efficient training and adaptation to smaller models.



\bibliography{custom}

\newpage
\appendix

\section{Experimental Setup}
\label{app:setup}

\subsection{LLMs Usage}
Large Language Models (LLMs) were used solely as general-purpose assistive tools to help polish the manuscript’s language and to refine instructions within our prompt templates. Specifically, LLMs aided in improving grammar, clarity, and style, and in suggesting alternative phrasings for prompt templates. All scientific ideas, experimental design, and key arguments were conceived and written by the authors, and all factual statements were independently verified.

\subsection{Datasets}
\label{app:dataset}

\subsubsection{MIMIC-IV-Ext Cardiac Dataset Details}
\label{app:mimic_dataset}

We use the MIMIC-IV-Ext Cardiac Disease dataset~\citep{PhysioNet-mimic-iv-ext-cardiac-disease-1.0.0}, a collection of de-identified ICU hospitalizations with structured clinical notes and gold-standard cardiac diagnoses. Each case includes the following note sections:

\begin{itemize}[nosep,leftmargin=*]
  \item \textbf{Chief complaint}: primary reason for admission (e.g., dyspnea, orthopnea).
  \item \textbf{History of present illness (HPI)}: narrative clinical history preceding presentation.
  \item \textbf{Physical exam}: vital signs and examination findings.
  \item \textbf{Imaging}: radiology reports (e.g., X-ray, CT, MRI).
  \item \textbf{Catheterization (CATH)}: invasive hemodynamic and procedural findings.
  \item \textbf{ECG / ECG machine report}: electrocardiographic interpretation.
  \item \textbf{Invasions}: invasive procedure documentation.
\end{itemize}

Ground-truth diagnoses are derived from ICD-10 codes with the cardiac prefix (``I''). To prevent label leakage and standardize note structure, we apply the following preprocessing steps:

\begin{enumerate}[nosep,leftmargin=*]
  \item Normalize line breaks across heterogeneous note formats.
  \item Remove placeholders and template artifacts (e.g., list markers, boilerplate fragments).
  \item Remove diagnostic summary sections (\texttt{IMPRESSION}, \texttt{FINAL DIAGNOSIS}) so the model must infer diagnoses from the clinical evidence.
  \item Deduplicate ECG machine text to eliminate repeated autogenerated phrases.
  \item Normalize whitespace by collapsing repeated spaces and empty lines.
\end{enumerate}


\subsubsection{BarExam Dataset Details}
\label{app:barexam_dataset}

We evaluate legal reasoning using the BarExam QA benchmark~\citep{zheng2025reasoning}, which consists of multiple-choice questions derived from the Multistate Bar Examination (MBE). Each instance includes:

\begin{itemize}[nosep,leftmargin=*]
  \item \textbf{Fact pattern}: narrative describing the legal scenario, including events, actors, and circumstances.
  \item \textbf{Question stem}: the specific legal issue to resolve.
  \item \textbf{Four answer choices} (A–D): one correct option and three distractors.
  \item \textbf{Gold legal passage}: an authoritative legal excerpt (statute, rule, or case law) supporting the correct answer.
\end{itemize}

Questions span six core MBE subjects: Constitutional Law, Contracts, Criminal Law, Evidence, Real Property, and Torts. The dataset draws from historical MBE administrations between 1972 and 1998.

\subsection{Retrieval Pipeline}
\label{app:retrieval}

Both domains use an optional retrieval pipeline to provide external evidence when required. We encode text using \texttt{all-MiniLM-L6-v2}~\citep{reimers2019sentence} (384-d embeddings) and index passages with FAISS \citep{douze2025faiss} for approximate nearest-neighbor search via cosine similarity. 

In the medical domain, we index cardiovascular clinical knowledge from the \texttt{ilyassacha/cardiologyChunks} dataset (3.2M records), chunked into 320-token segments with 64-token overlap, yielding 822{,}861 indexed chunks. Retrieval queries are constructed from the chief complaint and history of present illness, and the top-$k=3$ passages are inserted into the prompt as evidence.

In the legal domain, we index the BarExam legal passage corpus (856{,}835 passages) comprising case law paragraphs, Wex encyclopedia entries, and MBE explanations. Passages are pre-segmented and indexed directly. Depending on the experiment, models receive either the gold supporting passage or the top-$k=3$ retrieved passages using the question text as the retrieval query.

\begin{table}[h]
\centering
\small
\caption{Retrieval pipeline configuration.}
\label{tab:retrieval_config}
\begin{tabular}{@{}lll@{}}
\toprule
\textbf{Parameter} & \textbf{Medical} & \textbf{BarExam} \\
\midrule
Embedding model & \multicolumn{2}{c}{\texttt{all-MiniLM-L6-v2}} \\
Embedding dim. & \multicolumn{2}{c}{384} \\
Index type & \multicolumn{2}{c}{FAISS HNSW} \\
Chunk size & 320 tokens & N/A (pre-segmented) \\
Chunk overlap & 64 tokens & N/A \\
Top-$k$ & 3 & 1 (gold) / 3 (FAISS) \\
\bottomrule
\end{tabular}
\end{table}

\subsection{Model Backbones and Inference Configuration}
\label{app:models}
We evaluate EvidenceRL across multiple open-weight instruction-tuned model families spanning a range of parameter scales, including Llama~3.x, Gemma~3, and GPT-OSS variants. The evaluated models range from 3B to 120B parameters and represent diverse architectures and training recipes. Table~\ref{tab:models} lists all backbones used in our experiments.

\begin{table}[h]
\centering
\small
\caption{Model backbones used in experiments. All models run in bf16 via vLLM for inference.}
\label{tab:models}
\begin{tabular}{@{}llcc@{}}
\toprule
\textbf{Model} & \textbf{Family} & \textbf{Params} & \textbf{Inference} \\
\midrule
Llama-3.2-3B-It & Llama 3.2 & 3B & bf16, TP=2  \\
Llama-3.1-8B-It & Llama 3.1 & 8B & bf16, TP=2 \\
Llama-3.3-70B-It & Llama 3.3 & 70B & bf16, TP=4 \\
Gemma-3-4B-IT & Gemma 3 & 4B & bf16, TP=2  \\
Gemma-3-12B-IT & Gemma 3 & 12B & bf16, TP=2  \\
Gemma-3-27B-IT & Gemma 3 & 27B & bf16, TP=2 \\
GPT-OSS-20B & GPT-OSS & 20B & bf16, TP=2 \\
GPT-OSS-120B & GPT-OSS & 120B & bf16, TP=4  \\
\bottomrule
\end{tabular}
\end{table}

All inference is performed using \texttt{vLLM} with bfloat16 precision, which provides efficient batched decoding and tensor-parallel serving. Models up to 27B parameters use tensor parallelism with TP=2 across two NVIDIA H100 GPUs (80\,GB each); the 70B and 120B models use TP=4 across four NVIDIA H200 GPUs (141\,GB each). GPT-OSS generation uses a maximum generation length of 4{,}092 tokens. Detailed inference parameters are summarized in Table~\ref{tab:vllm_config}.

\begin{table}[h]
\centering
\small
\caption{vLLM inference configuration.}
\label{tab:vllm_config}
\resizebox{\linewidth}{!}{%
\begin{tabular}{@{}lll@{}}
\toprule
\textbf{Parameter} & \textbf{Standard} & \textbf{Self-Consistency} \\
\midrule
Temperature & 0.7 & 0.9 \\
Top-$p$ & 0.9 & 0.9 \\
Top-$k$ & 50 & 50 \\
Repetition penalty & 1.15 & 1.15 \\
Max tokens & 2{,}048 / 1{,}024$^*$ & 2{,}048 \\
$n$ (samples per prompt) & 1 & 10 \\
Tensor parallelism & 2--4 & 2 \\
GPU memory utilization & 0.90 & 0.90 \\
\bottomrule
\multicolumn{3}{l}{\footnotesize $^*$2{,}048 for medical, 1{,}024 for BarExam.} \\
\end{tabular}
}
\end{table}

\subsubsection{Self-Consistency Pipeline}
\label{app:sc}

Our self-consistency (SC) implementation follows \citet{wang2022self} but adapts majority voting for free-form clinical diagnosis text using embedding-based semantic clustering rather than exact-match voting.

\begin{enumerate}[nosep,leftmargin=*]
  \item \textbf{Diverse generation}: Generate $N=10$ completions per patient using higher temperature ($T=0.9$) via vLLM's batched $n$-parameter sampling.
  \item \textbf{Parse}: Extract 5 diagnoses from each completion (up to 50 diagnoses per patient).
  \item \textbf{Pool}: Collect all diagnosis names across all $N$ samples.
  \item \textbf{Cluster}: Embed all unique diagnosis names with \texttt{BioLORD-2023}~\citep{remy2024biolord} (768-dim, MPNet-based). Apply greedy agglomerative clustering: process names sorted by frequency (most common first); for each name, merge into the first existing cluster whose centroid has cosine similarity $\geq 0.85$, or start a new cluster.
  \item \textbf{Rank}: Rank clusters by a composite score: $\text{score} = \text{vote\_count} \times 100 - \text{avg\_position}$, where vote count is the number of distinct samples containing the diagnosis and avg\_position is its average rank across samples.
  \item \textbf{Select reasoning}: For the top-5 clusters, select the best reasoning paragraph (longest reasoning among votes ranked in top-2 within their respective samples).
\end{enumerate}

\subsubsection{Self-RAG Pipeline}
\label{app:self_rag}

Our Self-RAG \citep{asai2023self} implementation adapts the retrieve-then-read paradigm with an explicit self-critique step that makes retrieval \emph{adaptive}, only patients with uncertain diagnoses trigger evidence retrieval.

\begin{enumerate}[nosep,leftmargin=*]
  \item \textbf{Zero-shot generation}: Generate an initial set of 5 diagnoses with reasoning from patient context alone (no evidence).
  \item \textbf{Self-critique}: The same model evaluates each diagnosis, assigning a confidence level (\textsc{high}/\textsc{low}) and a boolean \texttt{needs\_evidence} flag. 
  \item \textbf{Conditional retrieval}: If any diagnosis is flagged as uncertain, a retrieval query is constructed from the uncertain diagnoses' names and reasoning. The top-$k{=}3$ chunks are retrieved via FAISS. Patients with all high-confidence diagnoses skip retrieval entirely.
  \item \textbf{Selective refinement}: For patients with retrieved evidence, the model regenerates all 5 diagnoses given the original patient context, its initial diagnoses (annotated with confidence tags), and the retrieved evidence. Patients without retrieval retain their zero-shot output.
\end{enumerate}

\section{Training and Reward Implementation}
\label{app:training_appendix}

\subsection{SFT, fDPO, and GRPO Datasets}
\label{app:training_datasets}

We construct separate training datasets for supervised fine-tuning (SFT), faithfulness DPO (fDPO), and GRPO reinforcement learning. All datasets are derived from the same 3{,}700 training patients in the MIMIC-IV-Ext cardiac dataset or 954 training cases across six legal subjects in BarExam.

\paragraph{SFT Dataset.}
SFT examples are generated from candidate outputs produced by all eight model backbones on the training cases. Each model predicts five ranked diagnoses with reasoning, yielding up to 29{,}600 candidate outputs. Outputs are filtered using the diagnostic taxonomy (Section~\ref{sec:metrics}). An output is retained if at least two of the top-3 diagnoses are \textit{Evidence-Based (EB)}, meaning the diagnosis is both correct and grounded ($r_g^{\max} > 0.5$). Qualified outputs are ranked by (i) the number of EB diagnoses in the top-3 and (ii) the mean $r_g^{\max}$ across the top-3 as a tiebreaker. After deduplication, we cap at two examples per patient to avoid over-representation. The final SFT dataset contains 4{,}534 examples from 2{,}634 patients.

\paragraph{fDPO Dataset.}
The fDPO dataset provides preference pairs that encourage grounded reasoning. For each training patient, all eight backbones generate responses under the no-retrieval setting. Each response is scored by the NLI grounding evaluator, and we compute the mean grounding score across the five predicted diagnoses. For each patient, the highest-scoring response is selected as the \textit{chosen} example and the lowest-scoring as the \textit{rejected} example. We apply the following filtering criteria:
\begin{enumerate}[nosep,leftmargin=*]
    \item grounding gap $\geq 0.7$ between chosen and rejected responses,
    \item chosen response grounding $\geq 0.1$,
    \item rejected response grounding $\leq -0.1$.
\end{enumerate}
This yields 2{,}292 preference pairs with a mean grounding gap of 1.30 (chosen mean\,=\,0.75, rejected mean\,=\,$-$0.55). No correctness filtering is applied, making fDPO a pure grounding preference objective.

\paragraph{GRPO Dataset.}
GRPO training uses all 3{,}700 training patients (954 in the legal domain) without filtering. Each entry contains the patient context, ground-truth diagnoses, and optional retrieved evidence (for RAG variants). Unlike SFT and fDPO, GRPO does not rely on pre-generated outputs; completions are generated on-policy during training and rewards are computed dynamically.

\subsection{Grounding NLI Models}

\paragraph{Medical Domain.}
We use PubMedBERT-MNLI-MedNLI\footnote{\url{https://huggingface.co/pritamdeka/PubMedBERT-MNLI-MedNLI}} as the frozen NLI model for grounding evaluation. This is a PubMedBERT-based~\citep{gu2021domain} cross-encoder fine-tuned on MultiNLI~\citep{williams2018broad} and MedNLI~\citep{romanov2018lessons}, providing domain-specific natural language inference for clinical text.

\paragraph{Legal Domain.}
We use nli-deberta-v3-large\footnote{\url{https://huggingface.co/cross-encoder/nli-deberta-v3-large}}, a DeBERTa-v3-Large cross-encoder fine-tuned on SNLI and MultiNLI. As no legal-domain NLI model is publicly available, we use this general-purpose model with token-aware premise truncation to ensure the hypothesis is never clipped at the 512-token input limit.

\subsection{Training Hyperparameters}

All models are fine-tuned using LoRA adapters~\citep{hu2022lora} applied to all transformer projection layers. 

\begin{table}[h]
\centering
\small
\caption{SFT training hyperparameters.}
\label{tab:sft_hparams}
\begin{tabular}{@{}ll@{}}
\toprule
\textbf{Parameter} & \textbf{Value} \\
\midrule
Epochs & 3 \\
Learning rate & $2 \times 10^{-5}$ \\
Scheduler & Cosine with 10\% warmup \\
Per-device batch size & 2 \\
Gradient accumulation & 8 \\
Max sequence length & 2{,}048 tokens \\
Loss masking & Completion-only \\
Precision & bfloat16 \\
\bottomrule
\end{tabular}
\end{table}

\begin{table}[h]
\centering
\small
\caption{fDPO training hyperparameters. Uses TRL \texttt{DPOTrainer} with sigmoid loss on faithfulness preference pairs.}
\label{tab:fdpo_hparams}
\begin{tabular}{@{}ll@{}}
\toprule
\textbf{Parameter} & \textbf{Value} \\
\midrule
Epochs & 2 \\
Learning rate & $5 \times 10^{-6}$ \\
Scheduler & Cosine with 10\% warmup \\
Per-device batch size & 1 \\
Gradient accumulation & 8 \\
DPO $\beta$ & 0.1 \\
Loss type & Sigmoid \\
Max sequence length & 8{,}192 tokens \\
Max prompt length & 6{,}144 tokens \\
Validation split & 5\% \\
Weight decay & 0.01 \\
Precision & bfloat16 \\
\bottomrule
\end{tabular}
\end{table}

\begin{table}[h]
\centering
\small
\caption{GRPO training hyperparameters. Implemented using TRL \texttt{GRPOTrainer} with vLLM colocate mode for on-policy generation.}
\label{tab:grpo_hparams}
\begin{tabular}{@{}ll@{}}
\toprule
\textbf{Parameter} & \textbf{Value} \\
\midrule
Epochs & 2 \\
Learning rate & $5 \times 10^{-6}$ \\
Scheduler & Cosine with 5\% warmup \\
Gradient accumulation & 8 \\
Group size $G$ & 8 completions \\
Temperature & 0.7 (medical); 0.7--1.0 (legal)\textsuperscript{$\dagger$} \\
Max prompt length & 6{,}144 tokens \\
Max completion length & 4{,}096 tokens \\
KL coefficient $\beta$ & 0.04 \\
Clipping $\epsilon$ & 0.2 \\
Policy updates per generation & 2 \\
Reward weights $(w_f, w_c, w_g)$ & (1, 1, 2) \\
Precision & bfloat16 \\
\bottomrule
\multicolumn{2}{l}{\footnotesize \textsuperscript{$\dagger$}Gemma use $T{=}1.0$ to increase exploration diversity.} \\
\end{tabular}
\end{table}

During GRPO, each of the $G{=}8$ completions per prompt is scored by three independent reward functions that share the same structure across domains but differ in their domain-specific implementations:

\begin{enumerate}[nosep,leftmargin=*]
  \item $R_{\text{format}}$: Binary format compliance. Returns $\{0, 1\}$.
  \begin{itemize}[nosep,leftmargin=*]
    \item \textit{Medical}: Valid JSON with exactly 5 diagnoses, each containing non-empty name and reasoning fields.
    \item \textit{Legal}: Valid JSON with an answer letter (A/B/C/D) and non-empty reasoning.
  \end{itemize}
  \item $R_{\text{correctness}}$: Task correctness. Returns $[0, 1]$.
  \begin{itemize}[nosep,leftmargin=*]
    \item \textit{Medical}: Embedding-based matching using \texttt{BioLORD-2023}~\citep{remy2024biolord}. For the top-3 predicted diagnoses, computes cosine similarity against all ground-truth diagnoses; a prediction is correct if max similarity exceeds $\tau{=}0.80$. Returns the fraction correct.
    \item \textit{Legal}: Exact match on the answer letter. Returns $\{0, 1\}$.
  \end{itemize}
  \item $R_{\text{grounding}}$: NLI-based evidence grounding, normalized to $[0, 1]$.
  \begin{itemize}[nosep,leftmargin=*]
    \item \textit{Medical}: Uses \texttt{PubMedBERT-MNLI-MedNLI}. Averages the max-grounding score across the top-3 diagnoses using the focus-then-verify architecture (Section~\ref{sec:reward}).
    \item \textit{Legal}: Uses \texttt{nli-deberta-v3-large}. Splits reasoning into sentences, scores each against the gold legal passage with token-aware premise truncation, and averages per-sentence scores.
  \end{itemize}
\end{enumerate}

\noindent

\begin{table}[h]
\centering
\small
\caption{LoRA adapter configuration (shared across SFT, fDPO, and GRPO).}

\label{tab:lora_config}
\begin{tabular}{@{}ll@{}}
\toprule
\textbf{Parameter} & \textbf{Value} \\
\midrule
Rank ($r$) & 16 \\
Alpha ($\alpha$) & 32 \\
Dropout & 0.05 \\
Target modules & \texttt{q\_proj, k\_proj, v\_proj, o\_proj,} \\
                & \texttt{gate\_proj, up\_proj, down\_proj} \\
Bias & None \\
Task type & \texttt{CAUSAL\_LM} \\
\bottomrule
\end{tabular}
\end{table}

\begin{figure*}
    \centering
    \includegraphics[width=\linewidth]{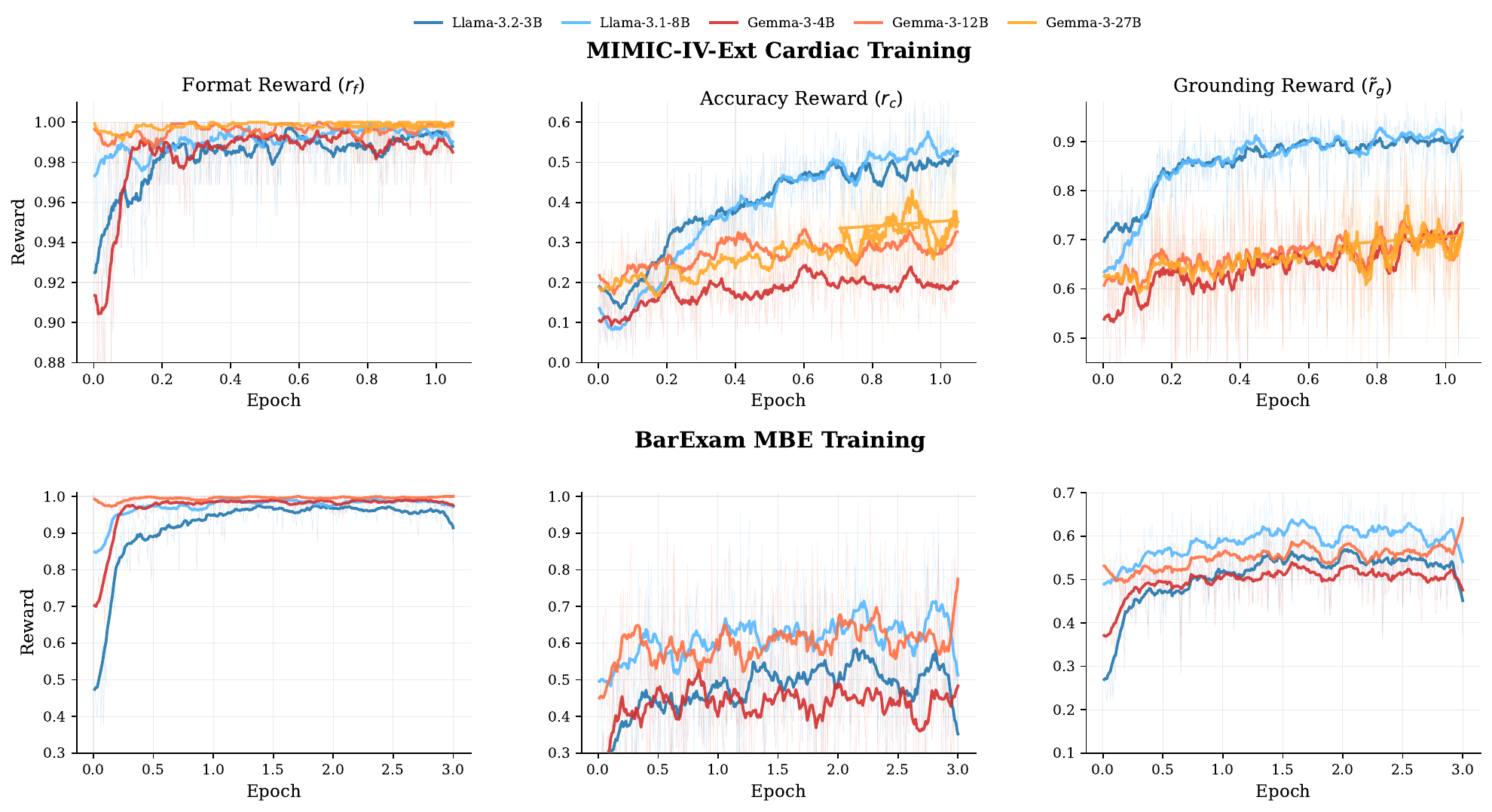}
    \caption{Training reward dynamics across model scales and objectives using MIMIC. We illustrate the training progress for the {Llama-3.1-8B}, {Llama-3.2-3B}, and the {Gemma-3} series (4B, 12B, and 27B) Performance is evaluated across three primary reward components: (left) {Format Reward ($r_{f}$)} measuring adherence to structural constraints; (center) Accuracy Reward ($r_{c}$) assessing the correctness of generated responses; and (right) Grounding Reward ($\tilde{r}_{g}$) quantifying the extent to which outputs are supported by provided context. Larger model scales generally exhibit higher reward ceilings and more stable convergence across all metrics.}
    \label{fig:training_curves}
\end{figure*}

Training curves (Figure~\ref{fig:training_curves}) reveal a consistent learning pattern across models. Format compliance converges quickly, with near-perfect structured outputs within the first few training steps. Accuracy reward improves rapidly during the first epoch, while grounding reward increases more gradually throughout training.

The largest improvements appear on the Llama models, while gains on Gemma models are more modest, suggesting that model families differ in how effectively reinforcement learning reshapes reasoning behavior. Llama models reach higher final reward values under GRPO, indicating that instruction-tuned models may retain additional capacity that reinforcement learning can unlock through targeted reward signals.

Overall, the results reveal a consistent pattern: inference-time techniques (retrieval and self-consistency) can influence output selection but rarely change the model’s underlying reasoning behavior. EvidenceRL, in contrast, alters how models generate diagnoses. The taxonomy analysis shows fewer hallucinations, fewer lucky guesses, and substantially higher evidence-based reasoning rates, suggesting that reinforcement learning shifts models from parametric reasoning toward genuine use of the clinical evidence provided in the prompt.


\section{Extended Results on MIMIC-IV-Ext}
\label{app:extended}
This appendix provides additional analyses supporting the results presented in Section~\ref{sec:results}. We examine the trade-off between diagnostic accuracy and evidence grounding, analyze the distribution of prediction types under the diagnostic taxonomy, inspect grounding distributions across patients, present a representative clinical case study, and audit the behavior of the grounding reward.

\subsection{Accuracy--Grounding Trade-Off}
\begin{figure}[h]
\centering
\includegraphics[width=\columnwidth]{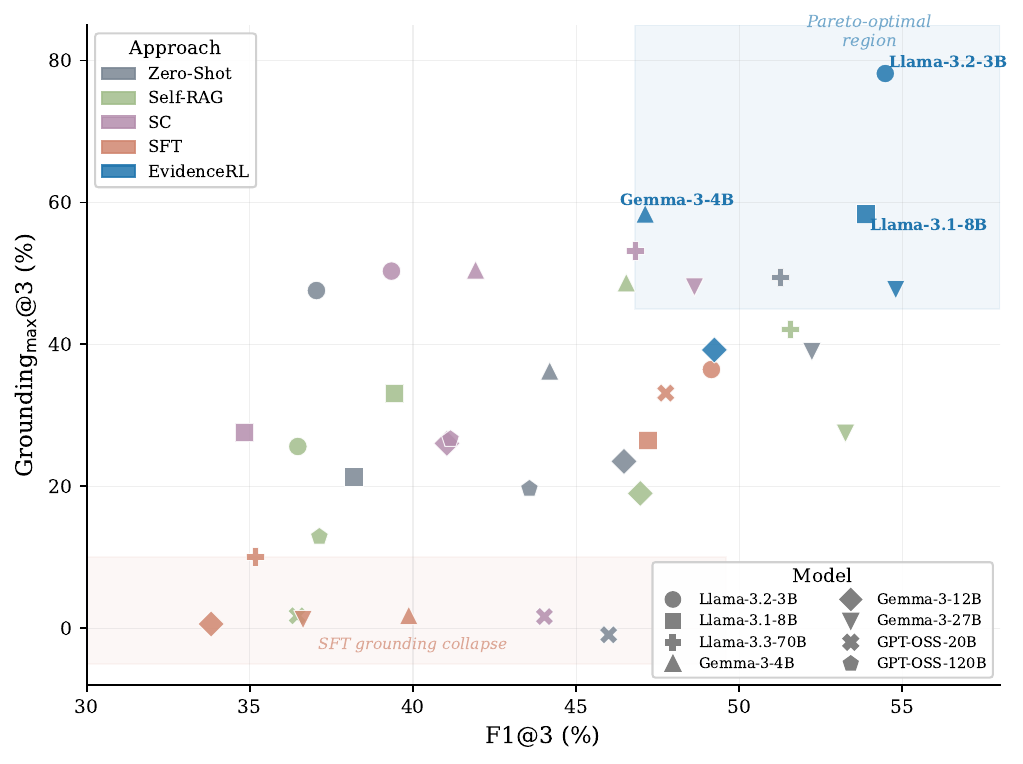}
\caption{F1@3 vs.\ $G_{\max}$@3 across all model backbones and approaches. EvidenceRL points (blue) occupy the Pareto-optimal region---high accuracy \emph{and} high grounding. SFT points (orange) collapse to near-zero grounding despite moderate accuracy. Self-RAG (green) clusters with or below zero-shot baselines.}
\label{fig:f1_vs_grounding}
\end{figure}

To better understand the interaction between diagnostic performance and evidential reliability, we plot model performance across the two axes of diagnostic accuracy (F1@3) and evidence grounding ($G_{\max}@3$). Figure~\ref{fig:f1_vs_grounding} shows the resulting Pareto landscape across all evaluated backbones and training methods.

EvidenceRL consistently occupies the Pareto-optimal region, achieving both higher diagnostic accuracy and stronger grounding than baseline approaches. In contrast, SFT models cluster in a region of moderate accuracy but near-zero grounding, indicating that the model learns to imitate plausible answers without grounding them in clinical evidence. Inference-time interventions such as Self-RAG and Self-Consistency remain near the zero-shot baseline, suggesting that retrieval or sampling alone does not reliably improve both axes simultaneously.

These results highlight a central property of EvidenceRL: by directly rewarding grounded correctness during training, the method shifts models toward solutions that jointly optimize accuracy and evidential reliability rather than trading one for the other.

\subsection{Taxonomy Analysis}
We further analyze model behavior using the diagnostic taxonomy introduced in Section~\ref{sec:metrics}, which categorizes predictions into Evidence-Based (EB), Weakly Supported (WS), Lucky Guess (LG), Hallucination (H), and Reasoning Failure (RF).

\begin{figure}[h]
\centering
\includegraphics[width=\columnwidth]{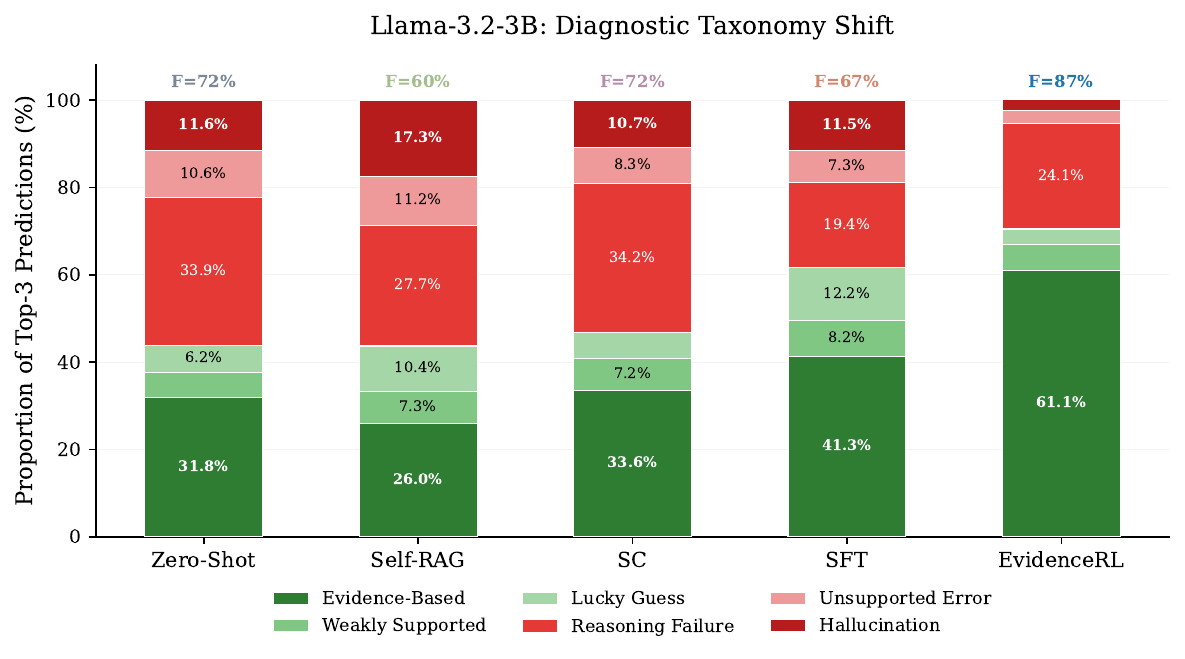}
\caption{Diagnostic taxonomy distribution for Llama-3.2-3B across all five approaches. Each bar decomposes top-3 predictions into six categories. Faithfulness (F) is annotated above each bar. EvidenceRL achieves the highest Evidence-Based rate (61.1\%) and Faithfulness (87\%), while Self-RAG has the lowest Faithfulness (60\%) due to elevated Lucky Guess and Hallucination rates.}
\label{fig:taxonomy_shift}
\end{figure}

Figure~\ref{fig:taxonomy_shift} shows the distribution of prediction types for Llama-3.2-3B across all evaluated approaches. EvidenceRL produces the largest proportion of Evidence-Based predictions, nearly doubling the EB rate relative to the zero-shot baseline. At the same time, hallucinations are substantially reduced.

In contrast, SFT improves the overall number of correct predictions but does not improve the evidential basis of those predictions. Much of its performance gain comes from increases in Lucky Guesses, indicating correct diagnoses that are unsupported by the evidence cited in the reasoning.

Inference-time interventions exhibit a different failure pattern. Self-RAG increases both hallucinations and lucky guesses, suggesting that retrieved documents can introduce confounding information when the model is not trained to critically evaluate evidence.

\subsection{Per-Patient Grounding Distributions}
\begin{figure}[!h]
\centering
\includegraphics[width=\columnwidth]{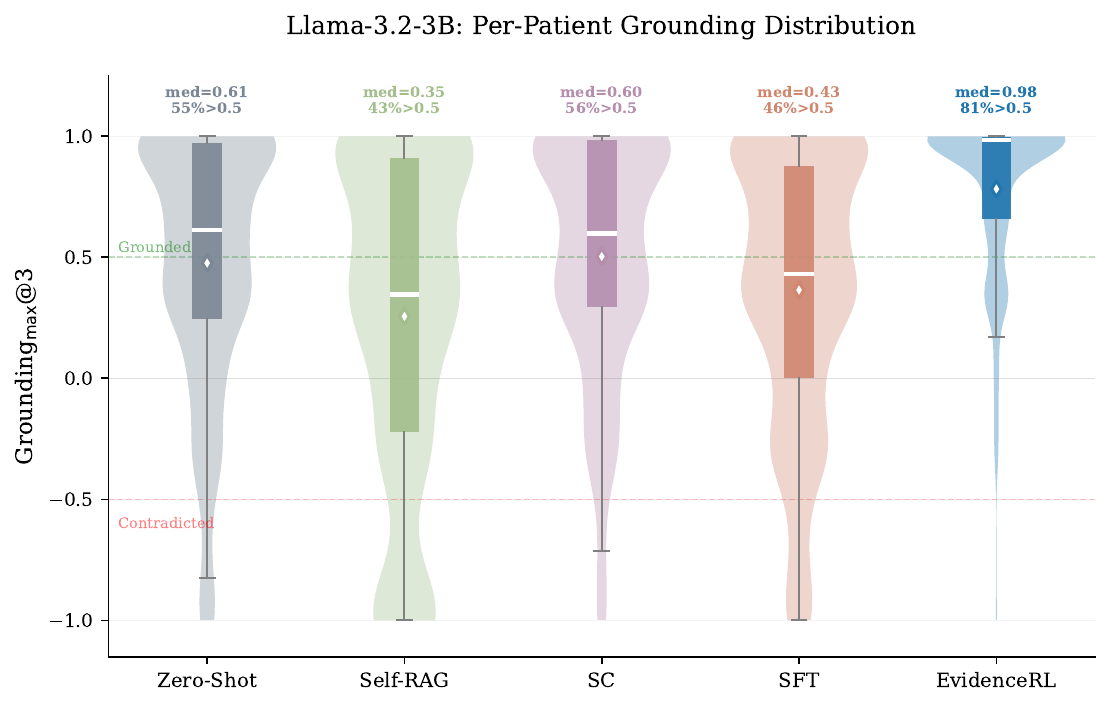}
\caption{Per-patient grounding distributions (Llama-3.2-3B). Each violin shows the full distribution of $G_{\max}$@3 across patients. Diamonds indicate means; white lines indicate medians. EvidenceRL concentrates mass near 1.0 (median=0.98, 81\% of patients above the grounding threshold), while other approaches show broad, bimodal distributions.}
\label{fig:grounding_violins}
\end{figure}
Aggregate grounding scores can obscure important differences in reliability across individual patients. To examine this, we plot the distribution of per-patient grounding scores ($G_{\max}@3$) for each approach (Figure~\ref{fig:grounding_violins}).

EvidenceRL produces a markedly tighter distribution, with the majority of patients exhibiting strong grounding scores. The median grounding score approaches the upper end of the scale, and a large fraction of patients exceed the grounding threshold used in our taxonomy.

Baseline approaches show much broader distributions. Zero-shot reasoning and Self-Consistency exhibit bimodal patterns, where grounding is strong for some patients but weak or contradictory for others. Self-RAG displays the most variability, reflecting sensitivity to retrieval quality: when retrieved passages are relevant grounding improves, but when retrieval is noisy grounding deteriorates substantially.

These results suggest that EvidenceRL not only improves average grounding scores but also stabilizes evidence use across patients, an important property for clinical decision support.
\begin{table*}[!t]
\centering
\small
\caption{Reasoning comparison for the top-ranked diagnosis of patient \#29017807 across all five approaches. Clinical values cited from the patient record are \underline{underlined}.}
\label{tab:case_study_reasoning}
\begin{tabular}{@{}p{1.5cm} p{2.2cm} c p{8.5cm}@{}}
\toprule
\textbf{Approach} & \textbf{Top-1 Dx} & \textbf{$G_{\max}$} & \textbf{Reasoning (excerpted)} \\
\midrule
EvidenceRL &
Myocardial Ischemia &
\textbf{+0.93} &
``The presence of a significant LAD lesion with \underline{90\% proximal occlusion}, combined with the patient's recent stress test showing \underline{antero-apical perfusion defect}, strongly suggests intermittent myocardial ischemia. The patient's \underline{elevated troponin level (4.67)} and \underline{reduced ejection fraction (LVEF = 42\%)} support the diagnosis.'' \\
\midrule
SFT &
NSTEMI &
$-$0.35 &
``The initial presentation included \underline{elevated Troponins (4.67 ng/mL)}, indicative of myocardial necrosis\ldots a previously documented stress test showing an \underline{apical perfusion defect} followed by development of apical thrombi\ldots warranting immediate evaluation.'' \\
\midrule
Zero-Shot &
CAD &
+0.14 &
``The patient presents with a history of NSTEMI and subsequent stress test showing \underline{ant/apical perfusion defects}\ldots The presence of \underline{90\% LAD lesion, 50\% CX}, and successful stenting of the LAD support this diagnosis.'' \\
\midrule
Self-RAG &
CAD &
$-$0.34 &
``The presence of a \underline{90\% proximal lesion in the LAD} explains the patient's angina\ldots The physical exam shows \underline{elevated blood pressure (SBP 153 mmHg)}\ldots Criteria from Evidence 1 and 2 support the presence of coronary artery disease.'' \\
\midrule
SC &
CAD &
+0.08 &
(Consensus from $N$=10 samples; reasoning from the highest-ranked generation.) \\
\bottomrule
\end{tabular}
\end{table*}

\subsection{Case Study: Patient \#29017807}
\label{app:case_study}

To illustrate how different approaches reason about the same patient, we present a detailed case study. Table~\ref{tab:case_study_summary} summarizes diagnostic performance; Table~\ref{tab:case_study_reasoning} shows the actual reasoning text.

\paragraph{Clinical presentation.}
A male patient with diabetes mellitus and chronic kidney disease (baseline creatinine 2.0) presented with chest pain, dyspnea, orthopnea, and paroxysmal nocturnal dyspnea. Key findings: stress test with anteroapical perfusion defect; echocardiogram showing apical thrombi, LVEF 42--45\%, mild mitral regurgitation, mild pulmonary hypertension; troponin of 4.67; cardiac catheterization revealing 90\% proximal LAD lesion and 80\% distal LCx disease, with successful PCI of the LAD.

\paragraph{Ground truth.} (1) Acute myocardial infarction, (2) Heart failure, (3) Chronic ischemic heart disease, (4) Hypertensive chronic kidney disease.

\begin{table}[h]
\centering
\small
\caption{Case study summary for patient \#29017807 (Llama-3.2-3B).}
\label{tab:case_study_summary}
\begin{tabular}{@{}l c c r@{}}
\toprule
\textbf{Approach} & \textbf{Correct} & \textbf{$G_{\mathrm{avg}}$@5} & \textbf{Combined} \\
\midrule
EvidenceRL   & \textbf{5/5} & \textbf{+0.758} & \textbf{0.879} \\
SFT          & 4/5          & $-$0.869        & $-$0.034 \\
SC           & 4/5          & +0.084          & 0.442 \\
Zero-Shot    & 3/5          & $-$0.694        & $-$0.047 \\
Self-RAG     & 3/5          & $-$0.861        & $-$0.130 \\
\bottomrule
\end{tabular}
\end{table}

\section{Prompt Templates}
\label{app:prompts}

Both domains share a common prompting architecture: a domain-expert system role, structured JSON output, and domain-specific \textit{synthesis rules} that guide the model to ground its reasoning in the provided evidence. We present the full prompt templates below.

Despite operating in different domains, both prompt families enforce the same structural principles: (1)~a domain-expert persona, (2)~structured JSON output for reliable parsing, (3)~three synthesis rules that require explicit citation of evidence rather than generic summaries, and (4)~critical instructions constraining the model to begin its response with the opening brace. The RAG$\to$no-RAG adaptation follows a parallel pattern in both domains: the evidence-citing rule (\textit{Guideline Alignment} in medical, \textit{Authority Anchoring} in legal) is replaced with a knowledge-based alternative (\textit{Avoid generic summaries} and \textit{Rule Statement}, respectively).

\definecolor{ragblue}{HTML}{E8F0FE}
\definecolor{noraggreen}{HTML}{E6F4EA}
\definecolor{judgeamber}{HTML}{FEF7E0}
\definecolor{ragborder}{HTML}{4285F4}
\definecolor{noragborder}{HTML}{34A853}
\definecolor{judgeborder}{HTML}{F9AB00}


\subsubsection{Medical Domain (Cardiac Diagnosis)}

\begin{tcolorbox}[
  colback=ragblue, colframe=ragborder,
  title={\textbf{Medical RAG Prompt}},
  fonttitle=\small\sffamily, breakable, sharp corners=south
]
\small\ttfamily
You are an expert cardiology clinical assistant. Based on the patient information and retrieved clinical evidence, provide exactly 5 cardiac diagnoses ranked from most to least likely.

\medskip
For EACH diagnosis, you MUST provide:\\
1. The diagnosis name (concise clinical term)\\
2. A reasoning paragraph following these "Clinical Synthesis" rules:\\
\quad - Pathophysiological Link: Explain how the specific symptoms (e.g., dyspnea) are directly explained by the clinical findings (e.g., the mitral regurgitation seen on ultrasound).\\
\quad - Evidence Integration: Use exact values from the Physical Exam (BP, RR) and Imaging (LVEF, PA pressures) as "anchors" for your argument.\\
\quad - Guideline Alignment: Explicitly state which criteria from the Retrieved Clinical Evidence are met by this specific patient's data.

\medskip
IMPORTANT: Your response MUST be valid JSON in exactly this format:\\
\{\\
\quad "diagnoses": [\\
\quad\quad \{"name": "Diagnosis 1 name", "reasoning": "..."\},\\
\quad\quad ...\\
\quad\quad \{"name": "Diagnosis 5 name", "reasoning": "..."\}\\
\quad ]\\
\}

\medskip
Patient Information:\\
\textit{\{patient\_context\}}

\medskip
Retrieved Clinical Evidence:\\
\textit{[Evidence 1]: \{passage\_1\}}\\
\textit{[Evidence 2]: \{passage\_2\}}\\
\textit{[Evidence 3]: \{passage\_3\}}

\medskip
CRITICAL INSTRUCTIONS FOR YOUR RESPONSE:\\
- Begin your response IMMEDIATELY with the opening brace \{\\
- Do NOT include any thinking, explanation, preamble, or commentary before the JSON\\
- Do NOT show your reasoning process outside the JSON -- all reasoning goes in the "reasoning" fields\\
- Output ONLY valid JSON, nothing else\\
- Start your response with \{
\end{tcolorbox}

\begin{tcolorbox}[
  colback=noraggreen, colframe=noragborder,
  title={\textbf{Medical No-RAG Prompt}},
  fonttitle=\small\sffamily, breakable, sharp corners=south
]
\small\ttfamily
You are an expert cardiology clinical assistant. Based on the patient information below, provide exactly 5 cardiac diagnoses ranked from most to least likely.

\medskip
For EACH diagnosis, you MUST provide:\\
1. The diagnosis name (concise clinical term)\\
2. A reasoning paragraph following these "Clinical Synthesis" rules:\\
\quad - Pathophysiological Link: Explain how the specific symptoms (e.g., dyspnea) are directly explained by the clinical findings (e.g., the mitral regurgitation seen on ultrasound).\\
\quad - Evidence Integration: Use exact values from the Physical Exam (BP, RR) and Imaging (LVEF, PA pressures) as "anchors" for your argument.\\
\quad - Avoid generic summaries: Do not just list facts; explain the "why" behind the diagnosis using the patient's unique data.

\medskip
IMPORTANT: Your response MUST be valid JSON in exactly this format:\\
\{\\
\quad "diagnoses": [\\
\quad\quad \{"name": "Diagnosis 1 name", "reasoning": "..."\},\\
\quad\quad ...\\
\quad\quad \{"name": "Diagnosis 5 name", "reasoning": "..."\}\\
\quad ]\\
\}

\medskip
Patient Information:\\
\textit{\{patient\_context\}}

\medskip
CRITICAL INSTRUCTIONS FOR YOUR RESPONSE:\\
- Begin your response IMMEDIATELY with the opening brace \{\\
- Do NOT include any thinking, explanation, preamble, or commentary before the JSON\\
- Do NOT show your reasoning process outside the JSON -- all reasoning goes in the "reasoning" fields\\
- Output ONLY valid JSON, nothing else\\
- Start your response with \{
\end{tcolorbox}

\begin{tcolorbox}[
  colback=judgeamber, colframe=judgeborder,
  title={\textbf{Medical LLM-as-Judge Prompt}},
  fonttitle=\small\sffamily, sharp corners=south
]
\small\ttfamily
You are evaluating the diagnosis prediction of a clinical model.

\medskip
CANDIDATE ANSWER: "\textit{\{predicted\_diagnosis\}}"

\medskip
ACCEPTED GROUND TRUTHS:\\
- \textit{\{ground\_truth\_1\}}\\
- \textit{\{ground\_truth\_2\}}\\
- ...

\medskip
TASK: Does the CANDIDATE ANSWER semantically match ANY of the ACCEPTED GROUND TRUTHS?\\
Respond `TRUE' if the candidate refers to the same underlying clinical concept as any item in the list (allowing for synonyms, abbreviations, or minor wording differences).\\
Respond `FALSE' if it represents a different clinical concept, severity, or anatomical location.

\medskip
Verdict:
\end{tcolorbox}


\subsubsection{Legal Domain (BarExam QA)}

\begin{tcolorbox}[
  colback=ragblue, colframe=ragborder,
  title={\textbf{Legal RAG Prompt}},
  fonttitle=\small\sffamily, breakable, sharp corners=south
]
\small\ttfamily
You are an expert legal analyst taking the Multistate Bar Examination (MBE). Based on the question and retrieved legal authorities, select the best answer.

\medskip
For your answer, you MUST provide:\\
1. The answer letter (A, B, C, or D)\\
2. A reasoning paragraph following these "Legal Synthesis" rules:\\
\quad - Factual Application: Identify the key facts from the question and explain how they map to the legal rule or doctrine at issue.\\
\quad - Authority Anchoring: Cite specific language, holdings, or principles from the Retrieved Legal Authorities as direct support for your conclusion. Do NOT reason beyond what the authorities state.\\
\quad - Rule-to-Fact Bridge: Explicitly connect the legal rule from the authority to the specific facts given, showing why the rule applies (or does not apply) to each answer choice.\\
\quad - Avoid generic summaries: Do not just restate the rule; explain the "why" using the specific facts of the question.

\medskip
IMPORTANT: Your response MUST be valid JSON in exactly this format:\\
\{\\
\quad "answer": "X",\\
\quad "reasoning": "Detailed legal reasoning..."\\
\}

\medskip
Question:\\
\textit{\{question\_text\_with\_choices\}}

\medskip
Retrieved Legal Authorities:\\
\textit{[Legal Authority 1]: \{passage\_1\}}\\
\textit{[Legal Authority 2]: \{passage\_2\}}\\
\textit{[Legal Authority 3]: \{passage\_3\}}

\medskip
CRITICAL INSTRUCTIONS FOR YOUR RESPONSE:\\
- Begin your response IMMEDIATELY with the opening brace \{\\
- The "answer" field MUST be exactly one letter: A, B, C, or D\\
- All reasoning goes in the "reasoning" field\\
- Output ONLY valid JSON, nothing else\\
- Start your response with \{
\end{tcolorbox}

\begin{tcolorbox}[
  colback=noraggreen, colframe=noragborder,
  title={\textbf{Legal No-RAG Prompt}},
  fonttitle=\small\sffamily, breakable, sharp corners=south
]
\small\ttfamily
You are an expert legal analyst taking the Multistate Bar Examination (MBE). Based on your legal knowledge, select the best answer.

\medskip
For your answer, you MUST provide:\\
1. The answer letter (A, B, C, or D)\\
2. A reasoning paragraph following these "Legal Synthesis" rules:\\
\quad - Factual Application: Identify the key facts from the question and explain how they map to the legal rule or doctrine at issue.\\
\quad - Rule Statement: State the applicable legal rule or doctrine from your knowledge.\\
\quad - Rule-to-Fact Bridge: Explicitly connect the legal rule to the specific facts given, showing why the rule applies (or does not apply) to each answer choice.\\
\quad - Avoid generic summaries: Do not just restate the rule; explain the "why" using the specific facts of the question.

\medskip
IMPORTANT: Your response MUST be valid JSON in exactly this format:\\
\{\\
\quad "answer": "X",\\
\quad "reasoning": "Detailed legal reasoning..."\\
\}

\medskip
Question:\\
\textit{\{question\_text\_with\_choices\}}

\medskip
CRITICAL INSTRUCTIONS FOR YOUR RESPONSE:\\
- Begin your response IMMEDIATELY with the opening brace \{\\
- The "answer" field MUST be exactly one letter: A, B, C, or D\\
- All reasoning goes in the "reasoning" field\\
- Output ONLY valid JSON, nothing else\\
- Start your response with \{
\end{tcolorbox}

\medskip

\begin{tcolorbox}[
  colback=judgeamber!30, colframe=judgeborder!60,
  fontupper=\small,
  sharp corners=south
]
\textbf{Legal Accuracy Evaluation.}\quad Unlike the medical domain, BarExam accuracy is evaluated via \textbf{exact match} on the answer letter (A/B/C/D), making an LLM-as-judge unnecessary.
\end{tcolorbox}





\end{document}